\documentclass{article}

\usepackage[dblblindworkshop,final]{neurips_2025}

\workshoptitle{MATH-AI}

\usepackage[utf8]{inputenc} 
\usepackage[T1]{fontenc}    
\usepackage[colorlinks,urlcolor=teal]{hyperref}
\usepackage{url}            
\usepackage{booktabs}       
\usepackage{amsfonts}       
\usepackage{amssymb}        
\usepackage{nicefrac}       
\usepackage{microtype}      
\usepackage{xcolor}         

\usepackage{graphicx}
\usepackage{amsmath}
\usepackage{multirow}
\usepackage{lipsum}
\usepackage{pmboxdraw}
\usepackage{float}
\usepackage{adjustbox}
\usepackage[most]{tcolorbox}
\tcbuselibrary{listings, breakable, skins}

\tcbset{
    promptstyle/.style={
        enhanced,
        breakable,
        colback=blue!5,
        colframe=blue!40,
        arc=2pt,
        boxrule=0.8pt,
        left=8pt,
        right=8pt,
        top=8pt,
        bottom=8pt,
        fonttitle=\bfseries\color{blue!50!black},
    }
}

\title{FractalBench: Diagnosing Visual-Mathematical Reasoning Through Recursive Program Synthesis}

\author{
  Jan Ondras$^*$\\
  MIT \\
  \texttt{janko@mit.edu} \\
  \And
  Marek \v{S}uppa$^*$\\
  Comenius University in Bratislava, Cisco \\
  \texttt{marek@suppa.sk}
}

\newcommand{\FB}{\textit{FractalBench}}

\begin{document}
\maketitle\vspace{-8mm}
\begin{abstract}\vspace{-2mm}
Mathematical reasoning requires abstracting symbolic rules from visual patterns---inferring the infinite from the finite. We investigate whether multimodal AI systems possess this capability through \FB{}, a benchmark evaluating fractal program synthesis from images. Fractals provide ideal test cases: Iterated Function Systems with only a few contraction maps generate complex self-similar patterns through simple recursive rules, requiring models to bridge visual perception with mathematical abstraction. We evaluate four leading MLLMs---GPT-4o, Claude 3.7 Sonnet, Gemini 2.5 Flash, and Qwen 2.5-VL---on 12 canonical fractals. Models must generate executable Python code reproducing the fractal, enabling objective evaluation. Results reveal a striking disconnect: 76\% generate syntactically valid code but only 4\% capture mathematical structure. Success varies systematically---models handle geometric transformations (Koch curves: 17-21\%) but fail at branching recursion (trees: <2\%), revealing fundamental gaps in mathematical abstraction. \FB{} provides a contamination-resistant diagnostic for visual-mathematical reasoning and is available at \href{https://github.com/NaiveNeuron/FractalBench}{https://github.com/NaiveNeuron/FractalBench}

\end{abstract}\vspace{-10mm}
\begin{figure}[h!]
    \centering
    \includegraphics[width=0.9\textwidth,trim={0 0.45cm 0 0},clip]{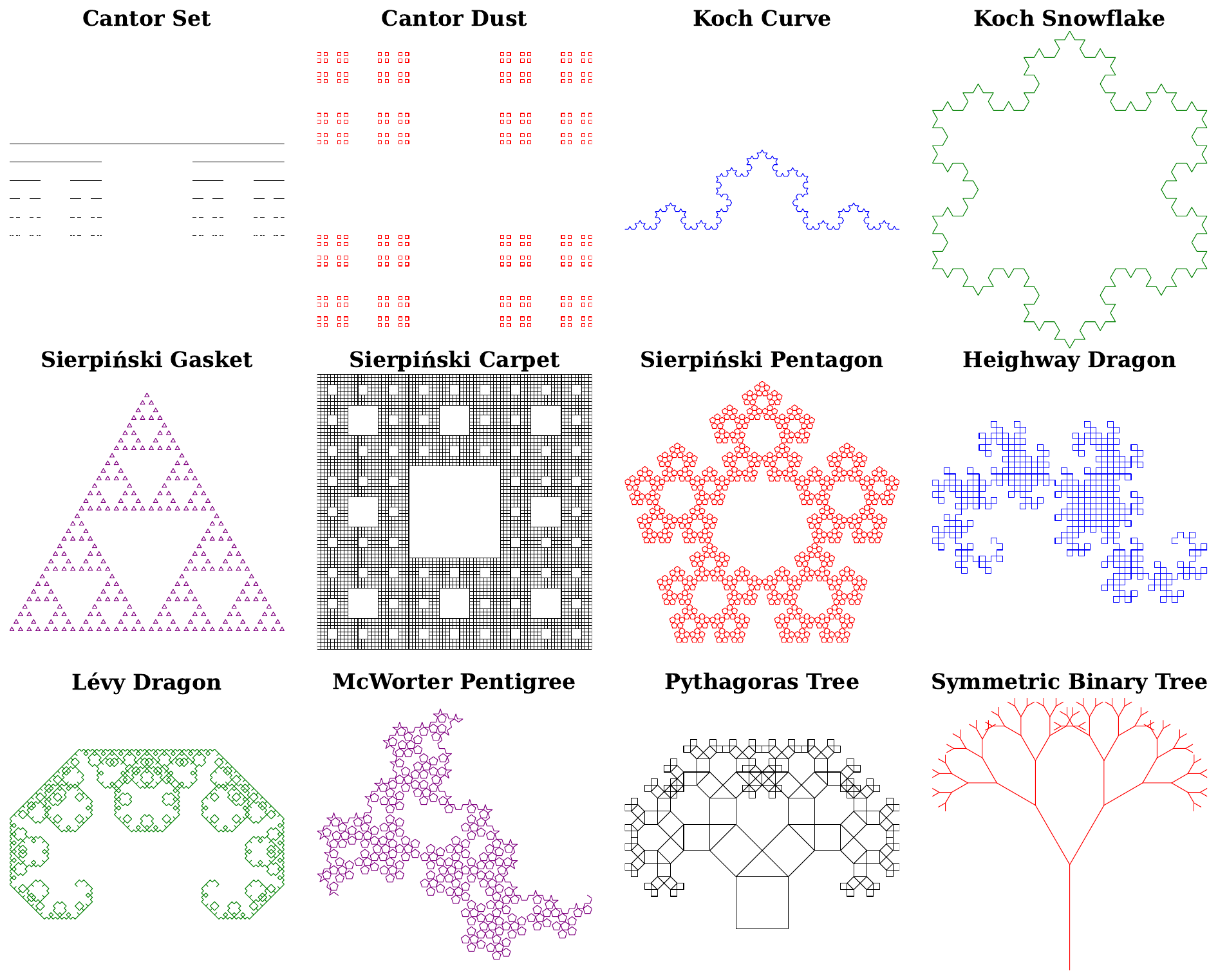}\vspace{-3mm}
    \caption{Twelve canonical fractals testing different mathematical reasoning capabilities: 
    linear recursion (Cantor),
    geometric transformations (Koch), 
    multi-scale self-similarity (Sierpiński), 
    space-filling curves (dragons), 
    and branching recursion (trees). 
    All defined via Iterated Function Systems.
    }
    \label{fig:fractals_collection}\vspace{-8mm}
\end{figure}

\section{Introduction}\vspace{-2mm}
From ancient geometry to modern AI, a central challenge in mathematical reasoning has been the ability to infer general principles from specific examples—to see the infinite process within finite observations. Can multimodal large language models (MLLMs) achieve this abstraction, inferring recursive symbolic programs from visual evidence alone?

Fractals provide an ideal testbed. Each fractal is compactly defined by an Iterated Function System (IFS)~\cite{barnsley2014fractals,mandelbrot1983fractal}---typically 2-8 contraction mappings---yet generates arbitrarily complex patterns through recursive self-similarity. Successfully synthesizing fractal code demands three interconnected capabilities: recognizing scale invariance across recursive levels, inferring precise geometric transformations from visual evidence, and achieving recursive abstraction—understanding the generative process rather than enumerating visible patterns.

We introduce \FB{}, comprising 12 canonical fractals spanning distinct challenges: Koch curves test geometric transformations, Sierpiński structures probe multi-scale self-similarity, dragon curves evaluate space-filling navigation, and tree fractals assess branching recursion. This diversity enables systematic diagnosis: which mathematical capabilities do current models possess, and where do they fail?

Evaluating four leading MLLMs on 7,320 fractal images (610 unique test images across 12 model-prompt combinations) reveals a striking disconnect: 76\% execution success but only 4\% visual correctness. Koch curves achieve 17-21\% accuracy, Sierpiński fractals 3-18\%, while tree fractals fail catastrophically at <2\%, revealing models can compose local operations but lack recursive abstraction.

This work makes three contributions to understanding visual-mathematical reasoning in AI systems. First, we establish a diagnostic framework connecting fractal synthesis to specific mathematical reasoning requirements, enabling systematic capability assessment. Second, we provide empirical evidence that current MLLMs possess geometric capabilities but fundamentally lack recursive abstraction—findings with implications for mathematical AI beyond fractals. Third, we demonstrate contamination-resistant evaluation through parameterizable complexity, offering a methodology applicable to future benchmarking efforts.

\vspace{-1mm}
\section{Related Work}\vspace{-2mm}
Existing benchmarks reveal gaps in visual-mathematical reasoning. \textbf{TurtleBench}~\cite{rismanchian-etal-2025-turtlebench} achieves only 19\% accuracy on simple geometric shapes, testing \textit{geometric perception}. \textbf{MathVista}~\cite{lu2023mathvista} and \textbf{MATH-Vision}~\cite{wang2024measuring} evaluate mathematical problem-solving with visual contexts, while \textbf{MATHGLANCE}~\cite{sun2024mathglance} reveals models \textit{``do not know where to look''} in mathematical diagrams. \textbf{GeoGramBench}~\cite{tian2024geogramben} targets geometric program reasoning, showing performance degradation with structural complexity. These benchmarks primarily test \textit{applying} mathematical knowledge to solve visual problems.

\FB{} tests \textit{mathematical abstraction}---inferring recursive generative rules from self-similar patterns. Where \textbf{TurtleBench}~\cite{rismanchian-etal-2025-turtlebench} asks ``can you draw what you see?'', \FB{} asks ``can you infer the infinite process generating finite observations?'' This capability—abstracting symbolic rules from visual examples—is central to mathematical discovery and reasoning. Fractals uniquely target this gap through recursive self-similarity, precise geometric transformations, and objective pixel-perfect evaluation. \FB{} is not a general vision or code benchmark, but rather a targeted diagnostic of visual-mathematical abstraction. Notably, the same difficulty patterns we observe (e.g., failures on branching recursion) appear in broader benchmarks such as \textbf{GeoGramBench}~\cite{tian2024geogramben}, \textbf{MathVista}~\cite{lu2023mathvista}, and \textbf{MATHGLANCE}~\cite{sun2024mathglance}, indicating that the limitations we expose are conceptual rather than API-specific. See App.~\ref{sec:app:related_work} for a comprehensive survey.

\vspace{-1mm}
\section{FractalBench}\vspace{-2mm}
\subsection{Fractal Definitions via Iterated Function Systems}\vspace{-1mm}
Self-similar fractals are defined as attractors of contractive Iterated Function Systems (IFS). Given contraction mappings $f_1,\dots,f_m : \mathbb{R}^d \to \mathbb{R}^d$, the IFS attractor is the unique compact set $K$ satisfying $K = \bigcup_{i=1}^m f_i(K)$. We consider 12 classic fractals~\cite{barnsley2014fractals,mandelbrot1983fractal} (Fig.~\ref{fig:fractals_collection}), spanning Cantor sets, Koch curves, Sierpiński structures, dragon curves, and tree fractals. Each has contraction ratio $r \in (0,1)$ determining scale reduction per iteration. Complete definitions are provided in App.~\ref{sec:app:fractal_defs}, and parameters in App.~\ref{sec:app:fractal_params}.

\subsection{Benchmark Design and Mathematical Reasoning Requirements}\vspace{-1.5mm}
\FB{} comprises 610 images ($1,024 \times 1,024$ pixels, 4-12 recursive levels) across 12 canonical fractals spanning distinct challenges: Cantor sets (recursive subdivision), Koch curves (geometric transformations), Sierpiński structures (multi-scale self-similarity), dragon curves (space-filling navigation), and trees (branching recursion). Variable depths and colors create difficulty gradients enabling contamination-resistant evaluation (App.~\ref{sec:app:test_set_config}). Color variants specifically prevent pretrained MLLMs from relying on cached visual embeddings of canonical black fractals, ensuring genuine visual-mathematical reasoning rather than memorized patterns.

Our evaluation employs a four-command MinimalTurtle interface (move, turn, pen up/down)---intentional by design, not a limitation. While allowing generation of arbitrary fractal patterns, this minimal interface isolates visual-to-symbolic rule abstraction from confounds such as library recall, memorized syntax, or heuristic shape primitives. Fractals can be fully expressed via move and turn operations; richer APIs (e.g., L-systems, matplotlib) would allow models to bypass mathematical reasoning through template recall. This constraint enhances diagnostic value, paralleling minimal-grammar reasoning tests in formal-language evaluation.

\label{sec:math_reasoning_req}Successfully synthesizing fractal code requires a hierarchy of five capabilities: \textbf{(1) Scale invariance recognition}---identifying that patterns repeat at different scales with specific contraction ratios (e.g., Sierpiński gasket's threefold self-similarity with $r=1/2$). \textbf{(2) Geometric transformation inference}---extracting precise rotation angles, scaling factors, and translations from visual evidence (Koch curve's $60^{\circ}$ rotations). \textbf{(3) Recursive structure abstraction}---understanding the generative process through self-referential rules rather than explicit enumeration. \textbf{(4) Compositional reasoning}---coordinating multiple recursive processes (Koch snowflake applies the same rule to three triangle edges). \textbf{(5) Branching recursion}---managing exponential computational complexity where each parent spawns multiple children with independent state. These progress from geometric operations through recursive abstraction to exponential branching, enabling systematic diagnosis of MLLM capabilities (detailed in App.~\ref{sec:app:detailed_math_req}).

\vspace{-2mm}
\section{Evaluation}\vspace{-2mm}
We evaluate MLLMs across three prompting strategies on 610 fractal images spanning 12 types, varying recursion depths, and colors. The evaluation pipeline encompasses three stages: generating code from multimodal prompts, executing it in a sandboxed environment with timeout protection, and assessing reconstruction quality through Jaccard Index similarity, a.k.a.~Intersection over Union~(IoU).

\vspace{-1mm}
\subsection{Experimental Methodology}\vspace{-1.5mm}
We evaluate four representative MLLMs: GPT-4o~\cite{gpt4o-openai-2024}, Claude 3.7 Sonnet~\cite{claude35sonnet-anthropic-2024}, Gemini 2.5 Flash~\cite{gemini20flash-google-2025}, and Qwen 2.5 VL 72B~\cite{qwen25vl72b-github-2025}. We employ three prompting strategies to probe different aspects of synthesis (App.~\ref{sec:app:prompts}). \textbf{Direct Code Generation (DCG)} tests raw image-to-code synthesis by directly mapping fractals to MinimalTurtle code without intermediate reasoning steps. In contrast, \textbf{Reasoning Then Code (RTC)} enforces structured analysis of fractal properties before code generation, requiring models to articulate their understanding explicitly. Finally, \textbf{Recursive Structure Focus (RSF)} emphasizes recursion as the central device, explicitly requiring base cases, self-similarity patterns, and parameter scaling in the generated code. All prompts include MinimalTurtle interface (App.~\ref{sec:app:minimal_turtle}) guide, examples, and formatting requirements.

Generated code executes in a sandboxed environment with a 30-second timeout, handling syntax errors, runtime exceptions, and non-terminating code. Successfully executed code generates $1,024 \times 1,024$ images using MinimalTurtle (App.~\ref{sec:app:minimal_turtle}). We evaluate reconstruction quality using Intersection over Union: $\text{IoU} = |\mathcal{B}_a \cap \mathcal{B}_m| / |\mathcal{B}_a \cup \mathcal{B}_m|$ where $\mathcal{B}_a$ and $\mathcal{B}_m$ are binary masks of ground truth and model-generated images. A 95\% similarity threshold defines correctness, consistent with vision-to-code benchmarks.

\vspace{-1mm}
\subsection{Results}\vspace{-1mm}
Our evaluation reveals a striking disconnect between syntactic and semantic capabilities. While 76.1\% of generated code executes successfully, only 4.2\% produces visually correct fractals (Tab.~\ref{tab:black_overview})—indicating syntactic competence without semantic understanding. Models generate valid Python producing visual output, but the wrong fractal, implementing \textit{some} recursive pattern without inferring the \textit{correct} generative rule.

\begin{table}[h!]
\centering
\caption{Performance Overview for Black Fractals:
Results across 3 prompt types and 4 models (1,464 total evaluations, 122 per condition).
Prompt Types: Direct Code Generation (DCG), Reasoning Then Code (RTC), Recursive Structure Focus (RSF).
Metrics: Run\% = execution success rate; Acc\% = visual correctness among runnable samples; Overall\% = end-to-end success rate.
}
\begin{adjustbox}{max width=0.9\textwidth}
\begin{tabular}{llrrrrr}
\toprule
\textbf{Prompt Type} & \textbf{Model} & \textbf{Runnable} & \textbf{Run\%} & \textbf{Correct} & \textbf{Acc\%} & \textbf{Overall\%} \\
\midrule
\multirow{4}{*}{\textbf{DCG}} & Claude 3.7 Sonnet & 100 & 82.0\% & 9 & 9.0\% & 7.4\% \\
 & Gemini 2.5 Flash & 29 & 23.8\% & 14 & \textbf{48.3\%} & \textbf{11.5\%} \\
 & GPT-4o & 115 & 94.3\% & 11 & 9.6\% & 9.0\% \\
 & Qwen 2.5-VL & 121 & 99.2\% & 4 & 3.3\% & 3.3\% \\
\midrule
\multirow{4}{*}{\textbf{RTC}} & Claude 3.7 Sonnet & 105 & 86.1\% & 3 & 2.9\% & 2.5\% \\
 & Gemini 2.5 Flash & 38 & 31.1\% & 4 & \textbf{10.5\%} & 3.3\% \\
 & GPT-4o & 118 & 96.7\% & 2 & 1.7\% & 1.6\% \\
 & Qwen 2.5-VL & 107 & 87.7\% & 6 & 5.6\% & \textbf{4.9\%} \\
\midrule
\multirow{4}{*}{\textbf{RSF}} & Claude 3.7 Sonnet & 106 & 86.9\% & 4 & \textbf{3.8\%} & \textbf{3.3\%} \\
 & Gemini 2.5 Flash & 35 & 28.7\% & 1 & 2.9\% & 0.8\% \\
 & GPT-4o & 120 & 98.4\% & 3 & 2.5\% & 2.5\% \\
 & Qwen 2.5-VL & 120 & 98.4\% & 0 & 0.0\% & 0.0\% \\
\midrule
\textbf{Overall Total} & & \textbf{1,114} & \textbf{76.1\%} & \textbf{61} & \textbf{5.5\%} & \textbf{4.2\%} \\
\bottomrule
\end{tabular}
\end{adjustbox}
\label{tab:black_overview}\vspace{-2mm}
\end{table}

As seen in Tab.~\ref{tab:fractal_type_analysis}, performance varies systematically with mathematical challenge. Koch fractals achieve the highest success rates (17-21\%), which we attribute to their reliance on \textit{iterative geometric transformations}—locally applying rotation, scaling, and translation operations. This success demonstrates that models can compose basic geometric operations. However, even here the 80\% failure rate reveals a crucial limitation: geometric intuition alone proves insufficient without true recursive abstraction. Sierpiński fractals achieve moderate performance (3-18\%) despite simpler structure. Models recognize visual \textit{similarity} but fail to infer precise \textit{scale invariance}—that the whole is composed of exact scaled copies with specific contraction ratios.
Tree fractals catastrophically fail (<2\%) despite having the simplest IFS definitions with only 2 maps. This failure isolates a specific bottleneck: \textit{branching recursion}, where single parents spawn multiple independent recursive children with separate state. Rather than implementing true branching, models substitute iterative loops or single-branch recursion, revealing their inability to represent exponentially growing tree-structured computation graphs. The contrast with linear recursion—where Cantor sets perform better—confirms that branching specifically, not recursion itself, constitutes the limiting factor.
See App.~\ref{sec:app:comprehensive_results} for comprehensive results and App.~\ref{sec:app:failure_cases} for failure cases.

\paragraph{Prompting Strategy Analysis.}\label{sec:prompting_strategy_analysis} Counterintuitively, direct code generation substantially outperforms reasoning-first approaches across Claude, Gemini, and GPT-4o (Tab.~\ref{tab:black_overview}: DCG 7.4-11.5\% vs. RTC 1.6-3.3\%, RSF 0.8-3.3\%), inverting the typical chain-of-thought advantage observed in mathematical reasoning tasks~\cite{wei2022chain}. This phenomenon suggests that \textit{verbose intermediate reasoning may interfere with precise visual-to-code synthesis}. We hypothesize three complementary explanations: First, explicit reasoning may anchor models on high-level descriptions (``the pattern branches at each level'') that are difficult to translate into exact geometric parameters (angles, scaling ratios, coordinates), creating a semantic gap between verbal analysis and numerical implementation. Second, reasoning-focused prompts increase output length and complexity, potentially exhausting attention mechanisms before reaching the critical code generation phase. Third, fractal synthesis requires tight visual-geometric coupling that direct image-to-code pathways may preserve better than verbally-mediated translation. This finding aligns with emerging evidence that structured prompting can constrain generative tasks requiring precise spatial or numerical outputs~\cite{liu2024mind,li2025structured}, contrasting with chain-of-thought benefits observed for high-level logical reasoning~\cite{wei2022chain}. 

\paragraph{Code Complexity.} Analysis in App.~\ref{sec:app:code_complexity} shows that Gemini produces more verbose code, recursion-focused prompts reduce complexity, while reasoning prompts show minimal impact. Notably, some fractals exhibit phase-transition behavior where code complexity initially grows with recursion depth but sharply drops once models recognize the recursive structure—revealing a threshold beyond which models shift from literal pixel-level descriptions to compressed algorithmic representations. This connects to algorithmic information theory: fractals have very low Kolmogorov complexity, making code length a proxy for whether models capture true structural compressibility.

\section{Conclusion}\vspace{-2mm}
Advancing mathematical reasoning in AI requires systems that can infer symbolic generative rules from visual patterns—bridging perception and abstraction. We introduced \FB{} as a diagnostic benchmark for this capability, using fractal program synthesis to isolate core competencies ranging from scale invariance recognition and geometric transformation inference through recursive abstraction and branching recursion. Evaluating 7,320 fractal images (610 unique test images across 12 model-prompt combinations) reveals current MLLMs possess geometric capabilities but lack recursive mathematical abstraction. The 76\% execution versus 4\% correctness disconnect demonstrates syntactic competence without semantic understanding. Models compose local operations successfully (Koch: 17-21\%) but fail at branching recursion (trees: <2\%), suggesting that their operation involves pattern matching rather than inferring generative processes.

\FB{} provides diagnostic insights beyond binary metrics, revealing \textit{which} mathematical reasoning capabilities current systems possess and \textit{where} they fail. As a contamination-resistant testbed, it enables measuring progress toward AI genuinely integrating visual perception with symbolic mathematical reasoning. These insights have implications for mathematical AI across domains: education systems requiring worked examples and explanations, formal verification tools needing program synthesis from specifications, and scientific discovery pipelines requiring abstraction from observational data. 

\section*{Limitations}

\paragraph{Evaluation Methodology.}
We perform a single generation per image and use binary IoU $\ge$ 95\% as the correctness criterion, following similar vision-to-code benchmarks such as TurtleBench~\cite{rismanchian-etal-2025-turtlebench}, which may not fully capture model stochasticity or assess how closely models capture underlying generative structure.
Complementary structure-aware metrics—such as branch count accuracy or recursive depth detection—could provide finer-grained diagnostics of which geometric properties models capture versus miss (see Appendix~\ref{sec:app:limitations_eval}).

\paragraph{Benchmark Scope.}
Our benchmark focuses on MLLMs rather than traditional program synthesis baselines~\cite{chaudhuri2025neurosymbolic,gulwani2017program}, and evaluates four representative models (GPT-4o, Claude 3.7 Sonnet, Gemini 2.5 Flash, Qwen 2.5-VL) without including recent reasoning-specialized systems such as OpenAI o1 or DeepSeek-R1.
Comparisons with specialized synthesis techniques would provide useful performance context, and evaluating newer reasoning-focused models would help determine whether \FB{} exposes universal failure modes or distinguishes genuinely stronger capabilities (see Appendix~\ref{sec:app:limitations_scope}).

\paragraph{Analysis Limitations.}
Our prompting strategy findings (Sec.~\ref{sec:prompting_strategy_analysis}) remain observational without stepwise ablations, and we do not demonstrate how benchmark diagnostics could guide targeted model improvements.
Controlled experiments isolating prompt complexity, reasoning depth, and instruction structure would enable causal interpretation, while establishing a feedback loop from diagnostics to improvements would validate \FB{} as a tool for advancing visual–mathematical reasoning research (see Appendix~\ref{sec:app:limitations_analysis}).

By releasing \FB{}, we invite the research community to address these limitations to enable more robust benchmarking of visual-mathematical reasoning in AI systems.

\begin{ack}
Authors contributed equally; author order is alphabetical.
Jan Ondras gratefully acknowledges support from G-Research through a \href{https://www.gresearch.com/nextgen/grants/}{research grant}.
This work has also been partially supported by grant APVV-21-0114.
\end{ack}

\bibliographystyle{unsrt}
\bibliography{refs}

\clearpage
\appendix

\section{Comprehensive Related Work Survey}\label{sec:app:related_work}

This appendix provides a detailed survey of related work across multiple research areas relevant to \FB{}, offering comprehensive context beyond the focused discussion in the main paper.

\subsection{Mathematical Reasoning in Multimodal Models}

The evaluation of mathematical reasoning capabilities in vision-language models has emerged as a critical research area. \textbf{MathVista}~\cite{lu2023mathvista} established the foundation for visual mathematical reasoning evaluation, revealing that even advanced models like GPT-4V fall short of human performance by significant margins. \textbf{MATH-Vision}~\cite{wang2024measuring} provides a comprehensive dataset of 3,040 high-quality mathematical problems with visual contexts, spanning 16 mathematical disciplines.

Recent specialized benchmarks have identified specific failure modes in mathematical reasoning. \textbf{ErrorRadar}~\cite{yan2024errorradar} focuses on error detection in mathematical problem-solving, while \textbf{EMMA}~\cite{hao2024emma} targets enhanced multimodal reasoning across mathematics, physics, chemistry, and coding domains. These evaluations consistently show that current models struggle with complex multi-step reasoning and visual-mathematical integration.

Notably, \textbf{MATHGLANCE}~\cite{sun2024mathglance} reveals that multimodal models \textit{``do not know where to look''} in mathematical diagrams, highlighting fundamental visual attention problems when processing geometric content---a challenge particularly relevant to fractal understanding.

\subsection{Geometric and Spatial Reasoning}

Spatial reasoning represents another dimension crucial to fractal understanding. \textbf{SpatialVLM}~\cite{chen2024spatialvlm} introduces comprehensive 3D spatial reasoning evaluation, while \textbf{SpatialEval}~\cite{wang2024picture} provides systematic assessment across relationship understanding, navigation, and counting tasks. These works consistently identify significant performance gaps between human and AI capabilities in spatial reasoning.

\textbf{GeoGramBench}~\cite{tian2024geogramben} specifically targets geometric program reasoning, revealing that models show clear performance degradation as structural complexity increases. This finding is particularly relevant to fractal synthesis, where recursive complexity scales exponentially with iteration depth.

\subsection{Program Synthesis and Evaluation Methodologies}

The broader program synthesis literature provides important context for evaluation methodologies. \textbf{HumanEval+}~\cite{liu2023your} demonstrates that traditional code evaluation metrics significantly overestimate model performance, while \textbf{CodeBLEU}~\cite{ren2020codebleu} attempts to address limitations of text-based similarity metrics through syntactic and semantic analysis.

Recent work on \textbf{visual program synthesis}~\cite{chen2024training} shows that while models can learn to generate code from visual specifications, they often fail to capture underlying algorithmic principles, instead relying on surface-level pattern matching.

\subsection{Fractal Generation and Mathematical Structures}

While extensive work exists on neural fractal generation for computer graphics~\cite{larsson2016fractalnet} and fractal-inspired neural architectures, the intersection of fractal understanding and code synthesis remains largely unexplored. Previous work has focused on generating fractals through neural networks rather than understanding fractal images to synthesize their generating algorithms.

Hsu et al.~\cite{hsu2025what} study the inverse problem of reasoning from images of abstract concepts to generative programs, demonstrating program synthesis as a bridge between visual patterns and algorithmic rules. However, their approach targets different structures like mazes rather than the mathematically rigorous recursive patterns of fractals.

\section{Fractal Definitions via Iterated Function Systems}\label{sec:app:fractal_defs}

Fractals can often be described as the attractors of \emph{Iterated Function Systems} (IFS), which are finite sets of contractive similarity maps $f_i : \mathbb{R}^d \to \mathbb{R}^d$ acting on Euclidean space, typically combining a scaling (contraction) with a rigid motion such as a rotation or translation. 
Below, we summarize twelve classical examples, grouped by their geometric type. In each case, the IFS is written explicitly, and we briefly describe the geometric intuition behind the construction. 
See Fig.~\ref{fig:fractals_collection} for the corresponding fractal images.

\subsection{Cantor-Type Fractals}
\begin{itemize}
    \item \textbf{Cantor Set.}
    The standard middle-third Cantor set is generated by repeatedly removing the open middle third from each interval. Equivalently, it is the attractor of the IFS on $[0,1]$ defined by
    $$
    f_1(x) = \tfrac{1}{3}x, 
    \quad 
    f_2(x) = \tfrac{1}{3}x + \tfrac{2}{3}.
    $$
    
    \item \textbf{Cantor Dust.}
    The Cartesian product $C \times C \subset \mathbb{R}^2$ of two Cantor sets, also called Cantor dust, is obtained from four similarity maps:
    $$
    f_{ij}(x,y) = \tfrac{1}{3}(x,y) + \left(\tfrac{2i}{3}, \tfrac{2j}{3}\right), 
    \quad i,j \in \{0,1\}.
    $$
    This construction generalizes naturally to higher dimensions, yielding fractal "dusts" supported on hypercubes.
\end{itemize}

\subsection{Koch-Type Fractals}
\begin{itemize}
    \item \textbf{Koch Curve.}
    The Koch curve is obtained by recursively replacing each line segment with four smaller segments forming an equilateral "bump". In the IFS form, it is defined by
    $$
    \begin{aligned}
    f_1(x) &= \tfrac{1}{3}x, \\
    f_2(x) &= \tfrac{1}{3}R_{\pi/3}(x) + \left(\tfrac{1}{3},0\right), \\
    f_3(x) &= \tfrac{1}{3}R_{-\pi/3}(x) + \left(\tfrac{1}{2},\tfrac{\sqrt{3}}{6}\right), \\
    f_4(x) &= \tfrac{1}{3}x + \left(\tfrac{2}{3},0\right),
    \end{aligned}
    $$
    where $R_{\theta}$ is a 2D rotation matrix corresponding to a counterclockwise rotation by angle $\theta$.
    
    \item \textbf{Koch Snowflake.}
    Applying the Koch curve construction simultaneously to each side of an equilateral triangle yields the famous Koch snowflake, a closed Jordan curve of infinite perimeter that nevertheless bounds a finite area.
\end{itemize}
    
\subsection{Sierpiński-Type Fractals}
\begin{itemize}
    \item \textbf{Sierpiński Gasket.}
    Starting from an equilateral triangle with vertices $v_1,v_2,v_3$, the gasket is the attractor of
    $$
    f_i(x) = \tfrac{1}{2}(x - v_i) + v_i, \quad i=1,2,3,
    $$
    which maps the entire triangle onto its three corner subtriangles.
    
    \item \textbf{Sierpiński Carpet.}
    Subdividing the unit square into a $3 \times 3$ grid and removing the central subsquare produces the Sierpiński carpet. The remaining eight squares correspond to the similarities
    $$
    f_i(x) = \tfrac{1}{3}x + t_i, \quad i=1,\dots,8,
    $$
    where $t_i$ is the translation vector that places the contracted square in the $i$-th position of the grid.
    
    \item \textbf{Sierpiński Pentagon.}
    Analogously, for a regular pentagon with vertices $v_1,\dots,v_5$, define
    $$
    f_i(x) = r(x - v_i) + v_i, \quad i=1,\dots,5,
    $$
    with contraction ratio $r = \tfrac{1}{1+\varphi} \approx 0.382$, where $\varphi = \tfrac{1 + \sqrt{5}}{2}$ is the Golden ratio. 
    The attractor is a fractal pentagonal analog of the gasket.
\end{itemize}

\subsection{Dragon Curves}
\begin{itemize}
    \item \textbf{Heighway Dragon.}
    The Heighway dragon, one of the most well-known fractal curves, is generated by
    $$
    f_1(x) = \tfrac{1}{\sqrt{2}} R_{\pi/4}(x), 
    \quad 
    f_2(x) = \tfrac{1}{\sqrt{2}} R_{-\,\pi/4}(x) + (1,0).
    $$
    Its attractor is a self-overlapping curve that fills a two-dimensional region of positive area.
    
    \item \textbf{Lévy Dragon (Lévy C Curve).}
    Closely related is the Lévy C curve, defined by
    $$
    f_1(x) = \tfrac{1}{\sqrt{2}} R_{\pi/4}(x), 
    \quad 
    f_2(x) = \tfrac{1}{\sqrt{2}} R_{-\,\pi/4}(x) + \left(\tfrac{1}{2}, \tfrac{1}{2}\right).
    $$
    Unlike the Heighway dragon, it forms a symmetric zig-zag curve without self-intersections.
\end{itemize}

\subsection{Tree Fractals}
\begin{itemize}
    \item \textbf{McWorter's Pentigree.}
    For a regular pentagon with vertices $v_i$, the system
    $$
    f_i(x) = r(x - v_i) + v_i, \quad i=1,\dots,5,
    $$
    with contraction ratio $r = \tfrac{1}{1+\varphi} \approx 0.382$ 
    produces a branching, tree-like structure with pentagonal symmetry.
    
    \item \textbf{Pythagoras Tree.}
    The Pythagoras tree is a branching fractal constructed from squares. 
    The process begins with a unit square. 
    At each step, two child squares of side length scaled by $1/\sqrt{2}$ are attached along the top edge of the parent square. One square is anchored in the top left corner of the parent and rotated counterclockwise by $\pi/4$, the other is anchored in the top right corner of the parent and rotated clockwise by $\pi/4$.
    Repeating this process indefinitely yields a branching, tree-like arrangement of squares.
    Formally, the IFS is given by
    $$
    f_1(x) = \tfrac{1}{\sqrt{2}} R_{\pi/4}(x) + t_1,
    \quad 
    f_2(x) = \tfrac{1}{\sqrt{2}} R_{-\pi/4}(x) + t_2,
    $$
    where $R_{\pm \pi/4}$ denote 2D rotation matrices, and the translation vectors $t_i$ position the rotated child squares precisely on the top edge of the parent square.
    
    \item \textbf{Symmetric Binary Tree.}
    In a general symmetric binary fractal tree, each parent segment spawns two scaled child copies rotated symmetrically by angles $\pm \theta$ about the parent's endpoint. 
    The corresponding IFS is
    $$
    f_1(x) = r R_{\theta}(x) + t_1, 
    \quad 
    f_2(x) = r R_{-\theta}(x) + t_2,
    $$
    where $r \in (0,1)$ is the custom contraction ratio, $R_{\pm \theta}$ are the 2D rotation matrices, and the translation vectors $t_i$ place the child segments at the tip of the parent segment.
\end{itemize}

\section{Fractal Parameters}\label{sec:app:fractal_params}
\begin{table}[H]
    \caption{Parameters of twelve classic fractals defined as attractors of Iterated Function Systems.
    $\varphi = \tfrac{1 + \sqrt{5}}{2} \approx 1.618$ is the Golden ratio.
    }
    \label{tab:fractal_params}
    \centering
    \begin{tabular}{lcc}
    \toprule
    \textbf{Fractal Type} & \textbf{\# Maps} & \textbf{Contraction Ratio} $r$ \\
    \midrule
    Cantor Set & 2 & $1/3$ \\
    Cantor Dust & 4 & $1/3$ \\
    Koch Curve & 4 & $1/3$ \\
    Koch Snowflake & 4 per edge & $1/3$ \\
    Sierpiński Gasket & 3 & $1/2$ \\
    Sierpiński Carpet & 8 & $1/3$ \\
    Sierpiński Pentagon & 5 & $\tfrac{1}{1+\varphi} \approx 0.382$ \\
    Heighway Dragon & 2 & $1/\sqrt{2}$ \\
    Lévy Dragon & 2 & $1/\sqrt{2}$ \\
    McWorter's Pentigree & 5 & $\tfrac{1}{1+\varphi} \approx 0.382$ \\
    Pythagoras Tree & 2 & $1/\sqrt{2}$ \\
    Symmetric Binary Tree & 2 & custom \\
    \bottomrule
    \end{tabular}
\end{table}

\section{Test Set Generation Configuration}\label{sec:app:test_set_config}

This section documents the complete configuration used to generate the \FB{} test set, enabling exact replication of our evaluation dataset.

\subsection{Rendering Parameters}

All fractal images are generated with the following specifications:
\begin{itemize}
    \item \textbf{Resolution:} 1,024 × 1,024 pixels
    \item \textbf{DPI:} 128 dots per inch
    \item \textbf{Line width:} 0.5 pixels
    \item \textbf{Background color:} White
    \item \textbf{Line colors:} Black, red, blue, green, purple (5 variations)
    \item \textbf{File format:} PNG with transparency support
\end{itemize}

\subsection{Fractal-Specific Configurations}
Each fractal is generated with fractal-specific parameters to ensure optimal visual quality and appropriate complexity scaling:
\begin{table}[H]
\caption{Fractal-Specific Generation Parameters: Images generated for all depths from 0 to the maximum depth, with initial base size of 500 pixels.
Additional Parameters list fractal-specific rendering options.}
\centering
\begin{adjustbox}{max width=\textwidth}
\begin{tabular}{lcc}
\toprule
\textbf{Fractal Type} & \textbf{Max Recursion Depth} & \textbf{Additional Parameters} \\
\midrule
Cantor Set & 5 & y\_spacing=20 \\
Cantor Dust & 4 & -- \\
Koch Curve & 4 & -- \\
Koch Snowflake & 5 & -- \\
Sierpiński Gasket & 6 & -- \\
Sierpiński Carpet & 4 & -- \\
Sierpiński Pentagon & 6 & -- \\
Heighway Dragon & 10 & -- \\
Lévy Dragon & 12 & -- \\
McWorter's Pentigree & 6 & -- \\
Pythagoras Tree & 8 & -- \\
Symmetric Binary Tree & 7 & angle=60°, ratio=0.65 \\
\bottomrule
\end{tabular}
\end{adjustbox}
\end{table}

\subsection{Maximum Recursion Depth}\label{sec:app:max_recursion_depth}

When rendering recursive fractals, the recursion depth must be capped so that substructures remain visually distinguishable. Continuing recursion beyond this limit would produce features smaller than the minimum resolvable size, which would likely lead to repeated images and thus duplicated samples in the dataset.

We compute the recursion depth limit by comparing the minimum distinguishable size to the contracted linear size of the initial base building block.
Let $s_0$ denote the initial linear size of the base building block in pixels (e.g., a single line segment, the first triangle, square, etc.) and let $r \in (0,1)$ denote the contraction ratio applied at each recursive step. At recursion depth $d$, the characteristic linear size of a substructure is approximately
$$
s_d \approx s_0 \cdot r^d.
$$
In particular, we take $s_0 = 500$ pixels and derive $r$ from fractal mathematical properties (Tab.~\ref{tab:fractal_params}) or, in the case of symmetric binary trees, specify explicitly, $r=0.65$.

To ensure that features remain resolvable, we impose the criterion
$$
s_d \geq s_{\min},
$$
where we set the minimum distinguishable size $s_{\min} = 1~\text{pixel}$.

Solving for the recursion depth yields
$$
d_{\max} = \left\lfloor \frac{\ln(s_{\min}/s_0)}{\ln(r)} \right\rfloor.
$$

For fractals with multiple contraction ratios (e.g., asymmetric branching), we would use the smallest ratio to avoid overestimating the resolvable depth. In practice, we also visually inspect the difference between the two highest admissible recursion depths to ensure that the last iteration still adds distinguishable structure.

\subsection{Dataset Structure}

The complete dataset organization follows this structure:
\begin{verbatim}
test_set/
├── black/
│   ├── cantor_set_depth0_size500_y_spacing20.png
│   ├── cantor_set_depth1_size500_y_spacing20.png
│   └── ... (progressive depth sequences)
├── red/
│   └── ... (same fractal sequences in red)
├── blue/
├── green/
└── purple/
\end{verbatim}

\textbf{File naming convention:} \texttt{<fractal\_name>\_<param><value>\_<fixed\_params>.png}

This systematic approach generates exactly 610 test images covering the complete parameter space of fractal types, complexity levels, and color variations, ensuring comprehensive evaluation coverage while maintaining consistent visual quality standards. Each color variant contains 122 images (610/5). We evaluate all images across 12 model–prompt combinations (4 models × 3 prompting strategies), yielding 1,464 runs per color and 7,320 runs in total across five colors.

\section{Detailed Mathematical Reasoning Requirements}\label{sec:app:detailed_math_req}

This appendix provides detailed explanations of the five mathematical reasoning capabilities required for fractal synthesis, expanding on the condensed framework in Sec.~\ref{sec:math_reasoning_req}.

\subsection{Scale Invariance Recognition}

Models must identify that visual patterns repeat identically at different scales, governed by a specific contraction ratio $r \in (0,1)$. For example, the Sierpiński gasket exhibits threefold self-similarity with $r = 1/2$, while the Cantor set uses $r = 1/3$. Recognizing scale invariance requires understanding that the fractal $K$ satisfies the self-similarity equation $K = \bigcup_{i=1}^m f_i(K)$ where each $f_i$ contracts by ratio $r$.

This capability goes beyond recognizing visual similarity---it requires inferring the precise quantitative relationship between scales. A model might visually recognize that a Sierpiński gasket contains three smaller copies of itself, but without determining that each copy is exactly half the size of the parent, it cannot generate correct code. Small errors in the contraction ratio accumulate through recursive iterations, causing the generated fractal to diverge from the target.

\subsection{Geometric Transformation Inference}

Beyond recognizing repetition, models must extract precise quantitative parameters---rotation angles, scaling factors, and translation vectors---from visual evidence alone. The Koch curve requires determining that edges rotate by exactly $\theta = \pi/3$ radians (60 degrees). Small errors accumulate exponentially through recursion, making this capability critical for correctness.

For instance, a model observing the Koch curve might recognize that line segments are subdivided and rotated, but must infer the exact angle from the visual pattern. An error of even 5 degrees compounds through recursion, producing a visually distinct fractal after a few iterations. This demands not just pattern recognition but quantitative inference from visual geometry.

\subsection{Recursive Structure Abstraction}

Models must understand the generative \textit{process} rather than merely enumerating visible patterns. This requires self-referential reasoning: recognizing that the whole is composed of transformed copies of itself. Implementing this demands recursive function calls, not iterative loops---a fundamental distinction in algorithmic thinking.

The key insight is understanding that the fractal is defined by the equation $K = \bigcup_i f_i(K)$---the set $K$ appears on both sides. This is fundamentally different from iterative construction where each level is explicitly computed from the previous one. Recursive abstraction means recognizing that the entire structure can be defined by a simple self-referential rule, rather than explicitly enumerating all levels. This is the essence of mathematical abstraction: finding the compact generative rule underlying complex patterns.

\subsection{Compositional Reasoning}

More complex fractals require composing multiple recursive processes. The Koch snowflake applies the Koch curve construction to three edges of a triangle simultaneously. The Pythagoras tree combines square placement with recursive branching. Models must coordinate multiple transformation rules operating at different structural levels.

Compositional reasoning tests whether models can maintain multiple simultaneous recursive processes with different parameters. For the Koch snowflake, this means applying the same recursive rule (Koch curve) to three different initial segments (the triangle edges), each with different orientations. For the Pythagoras tree, it means combining two distinct operations: positioning a square at a specific location and angle, then recursively applying the same process to the new squares. Failure in compositional reasoning manifests as applying the recursive rule to only one component or failing to maintain consistent parameters across multiple recursive calls.

\subsection{Branching Recursion}

Tree fractals introduce exponential complexity: a single parent spawns multiple recursive children, each requiring independent state maintenance. Unlike linear recursion (Cantor set) or fixed-arity iteration (Sierpiński structures), branching recursion demands representing and navigating tree-structured computation graphs.

The critical distinction is exponential growth in computational structure. In linear recursion (Cantor set), each recursive call spawns one child. In branching recursion (binary tree), each call spawns two independent children, creating $2^d$ nodes at depth $d$. This requires maintaining independent state for each branch, tracking position, angle, and scaling for each recursive child separately. Models consistently fail at this by: (1) using iterative loops that cannot maintain branching state, (2) implementing single-branch recursion that only follows one child, or (3) generating static approximations with fixed depth. True branching recursion requires representing and navigating tree-structured computation graphs---a capability current MLLMs fundamentally lack.

\section{Prompt Templates}\label{sec:app:prompts}

This section provides the complete prompt templates used in our evaluation, enabling reproducibility and detailed analysis of our methodology.

\subsection{Direct Code Generation Prompt}

\begin{tcolorbox}[promptstyle, title=Direct Code Generation Prompt]
Analyze the provided fractal image and write Python code that recreates it using MinimalTurtle.

\vspace{2mm}
Use MinimalTurtle to draw the fractal. You have access to:
\begin{itemize}
\item \texttt{turtle.move(distance)} -- move forward
\item \texttt{turtle.turn(degrees)} -- turn by angle
\item \texttt{turtle.goto(x, y)} -- move to position
\item \texttt{turtle.pen\_up()} / \texttt{turtle.pen\_down()} -- control drawing
\end{itemize}

\vspace{2mm}
Starter example:
\begin{lstlisting}[language=Python, basicstyle=\small\ttfamily, breaklines=true, frame=none, backgroundcolor=\color{gray!10}]
from utils.minimal_turtle import MinimalTurtle
from utils.minimal_renderer import render_turtle

def simple_line(length=100):
    turtle = MinimalTurtle()
    turtle.move(length)
    return turtle

if __name__ == "__main__":
    turtle = simple_line()
    render_turtle(turtle, "output.png")
\end{lstlisting}

\vspace{2mm}
Create a function that returns a configured MinimalTurtle instance and exposes any relevant parameters (e.g., depth, size, iterations).

\vspace{2mm}
Return your final answer in triple backticks with a complete, runnable implementation:
\begin{lstlisting}[language=Python, basicstyle=\small\ttfamily, breaklines=true, frame=none, backgroundcolor=\color{gray!10}]
# Your fractal implementation here

if __name__ == "__main__":
    turtle = your_function_name()
    render_turtle(turtle, "output.png")
\end{lstlisting}

\vspace{2mm}
\textbf{IMPORTANT:} Always include the \texttt{if \_\_name\_\_ == "\_\_main\_\_"} block so the code can be executed directly.
\end{tcolorbox}

\subsection{Reasoning Then Code Prompt}

\begin{tcolorbox}[promptstyle, title=Reasoning Then Code Prompt]
Study the fractal image, then document your reasoning before generating code.

\vspace{2mm}
Address these questions:
\begin{enumerate}
\item What type of fractal is this (e.g., recursive subdivision, L-system)?
\item What is the base shape or starting configuration?
\item Which transformation rules or drawing steps repeat each iteration?
\item How does complexity evolve with each iteration?
\end{enumerate}

\vspace{2mm}
After the analysis, produce MinimalTurtle code.

\vspace{2mm}
Use MinimalTurtle to draw the fractal. You have access to:
\begin{itemize}
\item \texttt{turtle.move(distance)} -- move forward
\item \texttt{turtle.turn(degrees)} -- turn by angle
\item \texttt{turtle.goto(x, y)} -- move to position
\item \texttt{turtle.pen\_up()} / \texttt{turtle.pen\_down()} -- control drawing
\end{itemize}

\vspace{2mm}
Reference snippet:
\begin{lstlisting}[language=Python, basicstyle=\small\ttfamily, breaklines=true, frame=none, backgroundcolor=\color{gray!10}]
from utils.minimal_turtle import MinimalTurtle
from utils.minimal_renderer import render_turtle

def square(size=50):
    turtle = MinimalTurtle()
    for _ in range(4):
        turtle.move(size)
        turtle.turn(90)
    return turtle

if __name__ == "__main__":
    turtle = square()
    render_turtle(turtle, "output.png")
\end{lstlisting}

\vspace{2mm}
Structure the response as:\\
\textbf{Analysis:} $\langle$Your reasoning$\rangle$\\
\textbf{Code:} Fractal implementation with required imports and main block
\end{tcolorbox}

\subsection{Recursive Structure Focus Prompt}

\begin{tcolorbox}[promptstyle, title=Recursive Structure Focus Prompt]
This fractal exhibits recursion. Produce Python code that makes the recursive structure explicit.

\vspace{2mm}
Highlight in your code:
\begin{enumerate}
\item The termination condition for recursion.
\item The self-similar pattern in each recursive call.
\item How parameters (e.g., length, angle, position) evolve between calls.
\end{enumerate}

\vspace{2mm}
Use MinimalTurtle to draw the fractal. You have access to:
\begin{itemize}
\item \texttt{turtle.move(distance)} -- move forward
\item \texttt{turtle.turn(degrees)} -- turn by angle
\item \texttt{turtle.goto(x, y)} -- move to position
\item \texttt{turtle.pen\_up()} / \texttt{turtle.pen\_down()} -- control drawing
\end{itemize}

\vspace{2mm}
Reference pattern:
\begin{lstlisting}[language=Python, basicstyle=\small\ttfamily, breaklines=true, frame=none, backgroundcolor=\color{gray!10}]
from utils.minimal_turtle import MinimalTurtle
from utils.minimal_renderer import render_turtle

def tree(turtle, length, depth):
    if depth == 0:
        return
    turtle.move(length)
    turtle.turn(30)
    tree(turtle, length * 0.7, depth - 1)
    turtle.turn(-60)
    tree(turtle, length * 0.7, depth - 1)

def create_tree(depth=5, initial_length=100):
    turtle = MinimalTurtle()
    tree(turtle, initial_length, depth)
    return turtle

if __name__ == "__main__":
    turtle = create_tree()
    render_turtle(turtle, "output.png")
\end{lstlisting}

\vspace{2mm}
Return your final answer in triple backticks with a complete, runnable implementation:
\begin{lstlisting}[language=Python, basicstyle=\small\ttfamily, breaklines=true, frame=none, backgroundcolor=\color{gray!10}]
# Your fractal implementation here

if __name__ == "__main__":
    turtle = your_function_name()
    render_turtle(turtle, "output.png")
\end{lstlisting}

\vspace{2mm}
\textbf{IMPORTANT:} Always include the \texttt{if \_\_name\_\_ == "\_\_main\_\_"} block so the code can be executed directly.
\end{tcolorbox}

\section{MinimalTurtle Graphics Interface}\label{sec:app:minimal_turtle}

Our evaluation uses a custom MinimalTurtle graphics library designed for fractal synthesis. This lightweight implementation provides essential turtle graphics operations while maintaining simplicity for code generation tasks.

\subsection{Core Interface}

The MinimalTurtle class provides the following key methods:

\begin{itemize}
    \item \texttt{move(distance)}: Move forward by specified distance
    \item \texttt{turn(angle)}: Turn by angle in degrees (positive = counterclockwise)
    \item \texttt{pen\_up()}, \texttt{pen\_down()}: Control drawing state
    \item \texttt{goto(x, y)}: Move to position without drawing
    \item \texttt{reset()}: Reset turtle to origin and clear paths
\end{itemize}

\subsection{Implementation}

The turtle maintains internal state including position $(x, y)$, heading angle, pen state, and drawing paths. All movements use standard trigonometry with heading 0° pointing east and positive angles rotating counterclockwise. The \texttt{get\_paths()} method returns line segments for rendering.

\begin{verbatim}
class MinimalTurtle:
    def __init__(self, x=0.0, y=0.0, heading=0.0):
        self.x, self.y, self.heading = x, y, heading
        self.pen_is_down = True
        self.paths = [[]]

    def move(self, distance):
        radians = math.radians(self.heading)
        new_x = self.x + distance * math.cos(radians)
        new_y = self.y + distance * math.sin(radians)
        if self.pen_is_down:
            if not self.paths[-1]:
                self.paths[-1].append((self.x, self.y))
            self.paths[-1].append((new_x, new_y))
        self.x, self.y = new_x, new_y
\end{verbatim}

This interface balances simplicity for LLM code generation with sufficient functionality for complex fractal rendering. The path-based approach enables efficient vectorized rendering at arbitrary resolutions.

\section{Comprehensive Evaluation Results}\label{sec:app:comprehensive_results}\vspace{-3mm}
\subsection{Complete Performance Overview Across All Colors}\vspace{-3mm}
\begin{table}[H]\vspace{-2mm}
\centering
\caption{Comprehensive Performance Overview: Complete results across 3 prompt types, 4 models, and 5 colors (7,320 total evaluations, 122 per condition).
Prompt Types: Direct Code Generation (DCG), Reasoning Then Code (RTC), Recursive Structure Focus (RSF).
Run\% = execution success rate; Acc\% = visual correctness among runnable samples; Overall\% = end-to-end success rate.
}\vspace{-1mm}
\begin{adjustbox}{max width=0.83\textwidth}
\begin{tabular}{lllrrrrr}
\toprule
\textbf{Prompt Type} & \textbf{Model} & \textbf{Color} & \textbf{Runnable} & \textbf{Run\%} & \textbf{Correct} & \textbf{Acc\%} & \textbf{Overall\%} \\
\midrule
\multirow{20}{*}{\textbf{DCG}} & \multirow{5}{*}{Claude 3.7 Sonnet} & {\color{black}$\blacksquare$} Black & 100 & 82.0\% & 9 & 9.0\% & 7.4\% \\
 &  & {\color{blue}$\blacksquare$} Blue & 103 & 84.4\% & 12 & 11.7\% & 9.8\% \\
 &  & {\color{green}$\blacksquare$} Green & 101 & 82.8\% & 13 & \textbf{12.9\%} & \textbf{10.7\%} \\
 &  & {\color{purple}$\blacksquare$} Purple & 98 & 80.3\% & 11 & 11.2\% & 9.0\% \\
 &  & {\color{red}$\blacksquare$} Red & 99 & 81.1\% & 7 & 7.1\% & 5.7\% \\
\cmidrule(lr){2-8}
 & \multirow{5}{*}{Gemini 2.5 Flash} & {\color{black}$\blacksquare$} Black & 29 & 23.8\% & 14 & \textbf{48.3\%} & \textbf{11.5\%} \\
 &  & {\color{blue}$\blacksquare$} Blue & 29 & 23.8\% & 12 & 41.4\% & 9.8\% \\
 &  & {\color{green}$\blacksquare$} Green & 30 & 24.6\% & 11 & 36.7\% & 9.0\% \\
 &  & {\color{purple}$\blacksquare$} Purple & 20 & 16.4\% & 5 & 25.0\% & 4.1\% \\
 &  & {\color{red}$\blacksquare$} Red & 28 & 23.0\% & 12 & 42.9\% & 9.8\% \\
\cmidrule(lr){2-8}
 & \multirow{5}{*}{GPT-4o} & {\color{black}$\blacksquare$} Black & 115 & 94.3\% & 11 & \textbf{9.6\%} & \textbf{9.0\%} \\
 &  & {\color{blue}$\blacksquare$} Blue & 114 & 93.4\% & 7 & 6.1\% & 5.7\% \\
 &  & {\color{green}$\blacksquare$} Green & 115 & 94.3\% & 9 & 7.8\% & 7.4\% \\
 &  & {\color{purple}$\blacksquare$} Purple & 116 & 95.1\% & 6 & 5.2\% & 4.9\% \\
 &  & {\color{red}$\blacksquare$} Red & 119 & 97.5\% & 7 & 5.9\% & 5.7\% \\
\cmidrule(lr){2-8}
 & \multirow{5}{*}{Qwen 2.5-VL} & {\color{black}$\blacksquare$} Black & 121 & 99.2\% & 4 & 3.3\% & 3.3\% \\
 &  & {\color{blue}$\blacksquare$} Blue & 120 & 98.4\% & 11 & \textbf{9.2\%} & \textbf{9.0\%} \\
 &  & {\color{green}$\blacksquare$} Green & 121 & 99.2\% & 9 & 7.4\% & 7.4\% \\
 &  & {\color{purple}$\blacksquare$} Purple & 121 & 99.2\% & 8 & 6.6\% & 6.6\% \\
 &  & {\color{red}$\blacksquare$} Red & 122 & 100.0\% & 7 & 5.7\% & 5.7\% \\
\midrule
\multirow{20}{*}{\textbf{RTC}} & \multirow{5}{*}{Claude 3.7 Sonnet} & {\color{black}$\blacksquare$} Black & 105 & 86.1\% & 3 & 2.9\% & 2.5\% \\
 &  & {\color{blue}$\blacksquare$} Blue & 105 & 86.1\% & 6 & \textbf{5.7\%} & \textbf{4.9\%} \\
 &  & {\color{green}$\blacksquare$} Green & 105 & 86.1\% & 3 & 2.9\% & 2.5\% \\
 &  & {\color{purple}$\blacksquare$} Purple & 108 & 88.5\% & 3 & 2.8\% & 2.5\% \\
 &  & {\color{red}$\blacksquare$} Red & 106 & 86.9\% & 1 & 0.9\% & 0.8\% \\
\cmidrule(lr){2-8}
 & \multirow{5}{*}{Gemini 2.5 Flash} & {\color{black}$\blacksquare$} Black & 38 & 31.1\% & 4 & 10.5\% & 3.3\% \\
 &  & {\color{blue}$\blacksquare$} Blue & 47 & 38.5\% & 4 & 8.5\% & 3.3\% \\
 &  & {\color{green}$\blacksquare$} Green & 43 & 35.2\% & 2 & 4.7\% & 1.6\% \\
 &  & {\color{purple}$\blacksquare$} Purple & 44 & 36.1\% & 5 & 11.4\% & \textbf{4.1\%} \\
 &  & {\color{red}$\blacksquare$} Red & 32 & 26.2\% & 4 & \textbf{12.5\%} & 3.3\% \\
\cmidrule(lr){2-8}
 & \multirow{5}{*}{GPT-4o} & {\color{black}$\blacksquare$} Black & 118 & 96.7\% & 2 & 1.7\% & 1.6\% \\
 &  & {\color{blue}$\blacksquare$} Blue & 116 & 95.1\% & 4 & \textbf{3.4\%} & \textbf{3.3\%} \\
 &  & {\color{green}$\blacksquare$} Green & 118 & 96.7\% & 3 & 2.5\% & 2.5\% \\
 &  & {\color{purple}$\blacksquare$} Purple & 112 & 91.8\% & 3 & 2.7\% & 2.5\% \\
 &  & {\color{red}$\blacksquare$} Red & 117 & 95.9\% & 2 & 1.7\% & 1.6\% \\
\cmidrule(lr){2-8}
 & \multirow{5}{*}{Qwen 2.5-VL} & {\color{black}$\blacksquare$} Black & 107 & 87.7\% & 6 & 5.6\% & 4.9\% \\
 &  & {\color{blue}$\blacksquare$} Blue & 113 & 92.6\% & 9 & 8.0\% & 7.4\% \\
 &  & {\color{green}$\blacksquare$} Green & 113 & 92.6\% & 11 & \textbf{9.7\%} & \textbf{9.0\%} \\
 &  & {\color{purple}$\blacksquare$} Purple & 114 & 93.4\% & 11 & 9.6\% & 9.0\% \\
 &  & {\color{red}$\blacksquare$} Red & 109 & 89.3\% & 10 & 9.2\% & 8.2\% \\
\midrule
\multirow{20}{*}{\textbf{RSF}} & \multirow{5}{*}{Claude 3.7 Sonnet} & {\color{black}$\blacksquare$} Black & 106 & 86.9\% & 4 & \textbf{3.8\%} & \textbf{3.3\%} \\
 &  & {\color{blue}$\blacksquare$} Blue & 106 & 86.9\% & 4 & \textbf{3.8\%} & \textbf{3.3\%} \\
 &  & {\color{green}$\blacksquare$} Green & 102 & 83.6\% & 1 & 1.0\% & 0.8\% \\
 &  & {\color{purple}$\blacksquare$} Purple & 108 & 88.5\% & 2 & 1.9\% & 1.6\% \\
 &  & {\color{red}$\blacksquare$} Red & 110 & 90.2\% & 2 & 1.8\% & 1.6\% \\
\cmidrule(lr){2-8}
 & \multirow{5}{*}{Gemini 2.5 Flash} & {\color{black}$\blacksquare$} Black & 35 & 28.7\% & 1 & 2.9\% & 0.8\% \\
 &  & {\color{blue}$\blacksquare$} Blue & 43 & 35.2\% & 1 & 2.3\% & 0.8\% \\
 &  & {\color{green}$\blacksquare$} Green & 42 & 34.4\% & 2 & \textbf{4.8\%} & \textbf{1.6\%} \\
 &  & {\color{purple}$\blacksquare$} Purple & 43 & 35.2\% & 2 & 4.7\% & \textbf{1.6\%} \\
 &  & {\color{red}$\blacksquare$} Red & 39 & 32.0\% & 1 & 2.6\% & 0.8\% \\
\cmidrule(lr){2-8}
 & \multirow{5}{*}{GPT-4o} & {\color{black}$\blacksquare$} Black & 120 & 98.4\% & 3 & \textbf{2.5\%} & \textbf{2.5\%} \\
 &  & {\color{blue}$\blacksquare$} Blue & 122 & 100.0\% & 1 & 0.8\% & 0.8\% \\
 &  & {\color{green}$\blacksquare$} Green & 118 & 96.7\% & 1 & 0.8\% & 0.8\% \\
 &  & {\color{purple}$\blacksquare$} Purple & 122 & 100.0\% & 1 & 0.8\% & 0.8\% \\
 &  & {\color{red}$\blacksquare$} Red & 121 & 99.2\% & 1 & 0.8\% & 0.8\% \\
\cmidrule(lr){2-8}
 & \multirow{5}{*}{Qwen 2.5-VL} & {\color{black}$\blacksquare$} Black & 120 & 98.4\% & 0 & 0.0\% & 0.0\% \\
 &  & {\color{blue}$\blacksquare$} Blue & 119 & 97.5\% & 1 & \textbf{0.8\%} & \textbf{0.8\%} \\
 &  & {\color{green}$\blacksquare$} Green & 119 & 97.5\% & 0 & 0.0\% & 0.0\% \\
 &  & {\color{purple}$\blacksquare$} Purple & 121 & 99.2\% & 1 & \textbf{0.8\%} & \textbf{0.8\%} \\
 &  & {\color{red}$\blacksquare$} Red & 120 & 98.4\% & 0 & 0.0\% & 0.0\% \\
\midrule
\textbf{Overall Total} & & & \textbf{5,627} & \textbf{76.9\%} & \textbf{310} & \textbf{5.5\%} & \textbf{4.2\%} \\
\bottomrule
\end{tabular}
\end{adjustbox}
\label{tab:comprehensive_overview}
\end{table}

\subsection{Fractal Type Analysis}

\begin{table}[H]
\centering
\caption{Fractal Type Analysis: Performance aggregated by fractal type across all depth variations and parameters.
Shows which fractal categories are most successfully generated by AI models, revealing systematic difficulty hierarchies from Koch-type curves (highest accuracy) to tree fractals (lowest accuracy).
Metrics: 
IoU = Intersection over Union; Std Dev = standard deviation of IoU scores.}
\begin{tabular}{lrrrrr}
\toprule
\textbf{Fractal Type} & \textbf{Total} & \textbf{Correct} & \textbf{Accuracy} & \textbf{Mean IoU} & \textbf{Std Dev} \\
\midrule
Koch Snowflake & 331 & 69 & 20.8\% & 0.343 & 0.382 \\
Sierpiński Carpet & 297 & 55 & 18.5\% & 0.288 & 0.374 \\
Koch Curve & 314 & 54 & 17.2\% & 0.222 & 0.372 \\
Sierpiński Pentagon & 284 & 24 & 8.5\% & 0.133 & 0.271 \\
Cantor Dust & 296 & 19 & 6.4\% & 0.140 & 0.252 \\
McWorter's Pentigree & 312 & 15 & 4.8\% & 0.067 & 0.211 \\
Cantor Set & 295 & 11 & 3.7\% & 0.055 & 0.191 \\
Sierpiński Gasket & 450 & 14 & 3.1\% & 0.128 & 0.183 \\
Lévy Dragon & 831 & 16 & 1.9\% & 0.041 & 0.146 \\
Pythagoras Tree & 757 & 14 & 1.8\% & 0.040 & 0.135 \\
Heighway Dragon & 899 & 14 & 1.6\% & 0.041 & 0.133 \\
Symmetric Binary Tree & 561 & 5 & 0.9\% & 0.024 & 0.094 \\
\bottomrule
\end{tabular}
\label{tab:fractal_type_analysis}
\end{table}

\subsection{Color Variation Analysis}

\begin{table}[H]
\centering
\caption{Color Variation Analysis: Performance across 5 fractal colors, aggregated across all models and prompt types. 
Shows minimal impact of color complexity on generation accuracy and visual similarity, with blue fractals achieving marginally higher success rates.}
\begin{tabular}{lrrrrr}
\toprule
\textbf{Color} & \textbf{Total} & \textbf{Correct} & \textbf{Accuracy} & \textbf{Mean IoU} & \textbf{Median IoU} \\
\midrule
{\color{blue}$\blacksquare$} Blue & 1,137 & 72 & 6.3\% & 0.107 & 0.013 \\
{\color{green}$\blacksquare$} Green & 1,127 & 65 & 5.8\% & 0.100 & 0.012 \\
{\color{black}$\blacksquare$} Black & 1,114 & 61 & 5.5\% & 0.100 & 0.013 \\
{\color{purple}$\blacksquare$} Purple & 1,127 & 58 & 5.1\% & 0.097 & 0.013 \\
{\color{red}$\blacksquare$} Red & 1,122 & 54 & 4.8\% & 0.091 & 0.011 \\
\bottomrule
\end{tabular}
\label{tab:color_analysis}
\end{table}

\subsection{Model Performance Summary}

\begin{table}[H]
\centering
\caption{Model Performance Summary: Evaluation metrics for each model across all fractal generation tasks, showing total images generated, correct visual matches, accuracy percentages, and similarity score statistics.
Gemini 2.5 Flash achieves the highest accuracy.
IoU = Intersection over Union.
}
\begin{tabular}{lrrrrr}
\toprule
\textbf{Model} & \textbf{Total} & \textbf{Correct} & \textbf{Accuracy} & \textbf{Mean IoU} & \textbf{Median IoU} \\
\midrule
Gemini 2.5 Flash & 542 & 80 & 14.8\% & 0.204 & 0.020 \\
Claude 3.7 Sonnet & 1,562 & 81 & 5.2\% & 0.095 & 0.012 \\
Qwen 2.5-VL & 1,760 & 88 & 5.0\% & 0.082 & 0.011 \\
GPT-4o & 1,763 & 61 & 3.5\% & 0.088 & 0.013 \\
\bottomrule
\end{tabular}
\label{tab:model_summary}
\end{table}

\subsection{Statistical Significance Analysis}

\begin{table}[H]
\centering
\caption{Pairwise Model Comparison with Statistical Significance Analysis: 
Mann-Whitney U tests for model performance differences in accuracy and IoU (Intersection over Union) scores, with $p$-values and Sig.~columns indicating statistical significance.
Cohen's $d$ provides effect size estimation.
}
\begin{adjustbox}{max width=\textwidth}
\begin{tabular}{llrrrcc}
\toprule
\textbf{Model A} & \textbf{Model B} & \textbf{Acc $p$-value} & \textbf{IoU $p$-value} & \textbf{Cohen's $d$} & \textbf{Acc Sig.} & \textbf{IoU Sig.} \\
\midrule
Claude 3.7 Sonnet & Gemini 2.5 Flash & 0.0278 & 0.0004 & -0.982 & Yes & Yes \\
Claude 3.7 Sonnet & GPT-4o & 0.1247 & 0.0010 & 0.508 & No & Yes \\
Claude 3.7 Sonnet & Qwen 2.5-VL & 0.8033 & 0.6686 & 0.052 & No & No \\
Gemini 2.5 Flash & GPT-4o & 0.0021 & 0.1166 & 1.145 & Yes & No \\
Gemini 2.5 Flash & Qwen 2.5-VL & 0.0343 & 0.0008 & 1.002 & Yes & Yes \\
GPT-4o & Qwen 2.5-VL & 0.3722 & 0.0000 & -0.466 & No & Yes \\
\bottomrule
\end{tabular}
\end{adjustbox}
\label{tab:statistical_comparison}
\end{table}

\section{Representative Failure Cases}\label{sec:app:failure_cases}

This section presents representative examples of fractal generation failures where models produced visual output but failed to correctly implement the intended fractal structure. These examples demonstrate the challenges in translating fractal mathematical definitions into correct algorithmic implementations, showing cases where models understood basic drawing concepts but missed crucial recursive patterns.

\begin{figure}[H]
    \centering
    \begin{tabular}{ccc}
        \textbf{Generated} & \textbf{Ground Truth} & \textbf{Fractal Type} \vspace{1mm}\\
        \adjustbox{valign=c}{\includegraphics[width=0.25\textwidth]{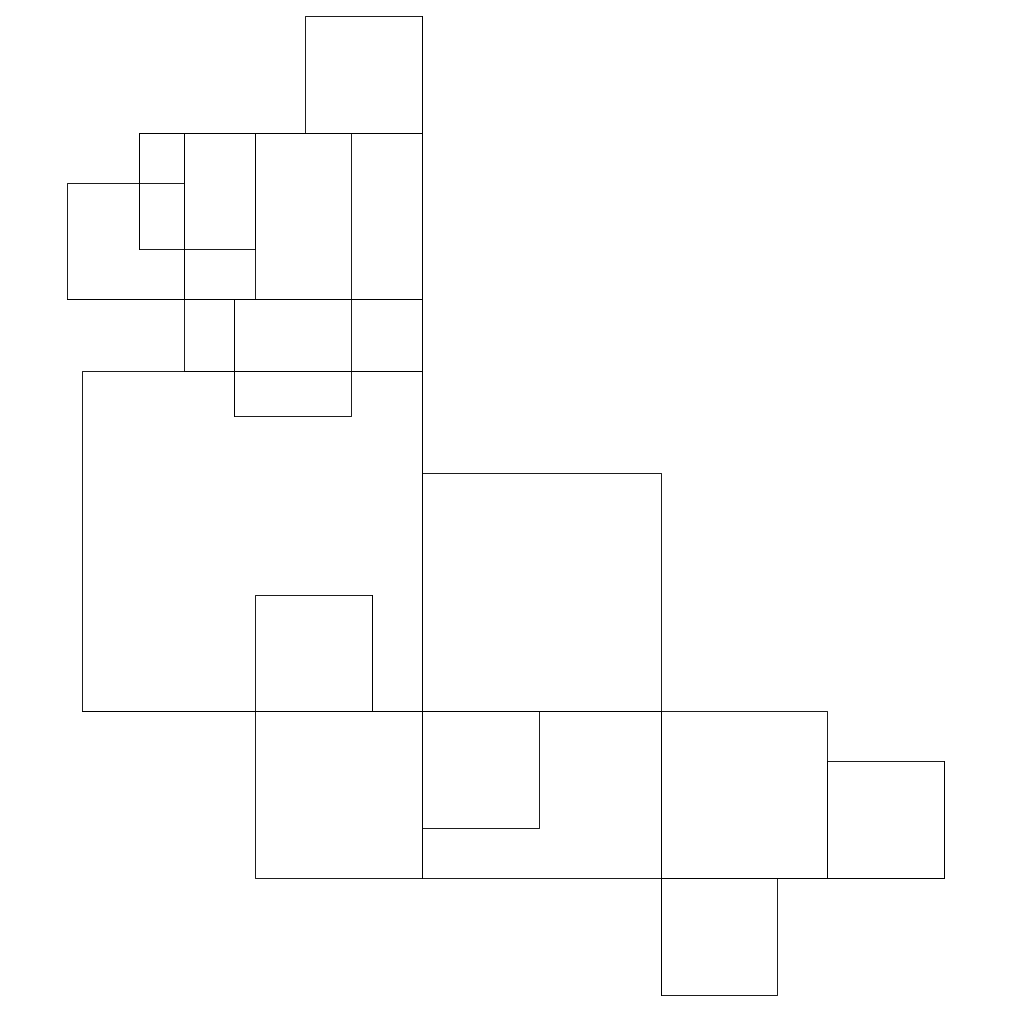}} &
        \adjustbox{valign=c}{\includegraphics[width=0.25\textwidth]{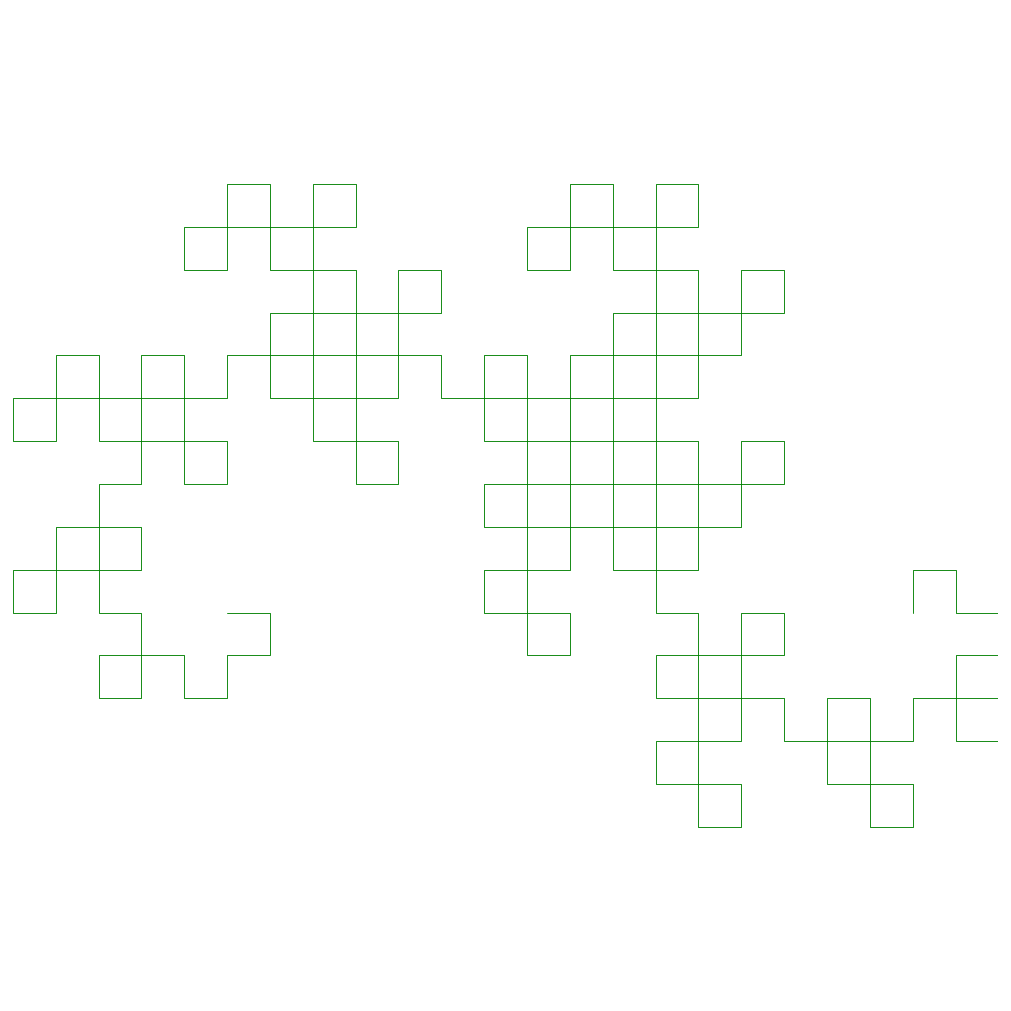}} & 
        Heighway Dragon \\

        \adjustbox{valign=c}{\includegraphics[width=0.25\textwidth]{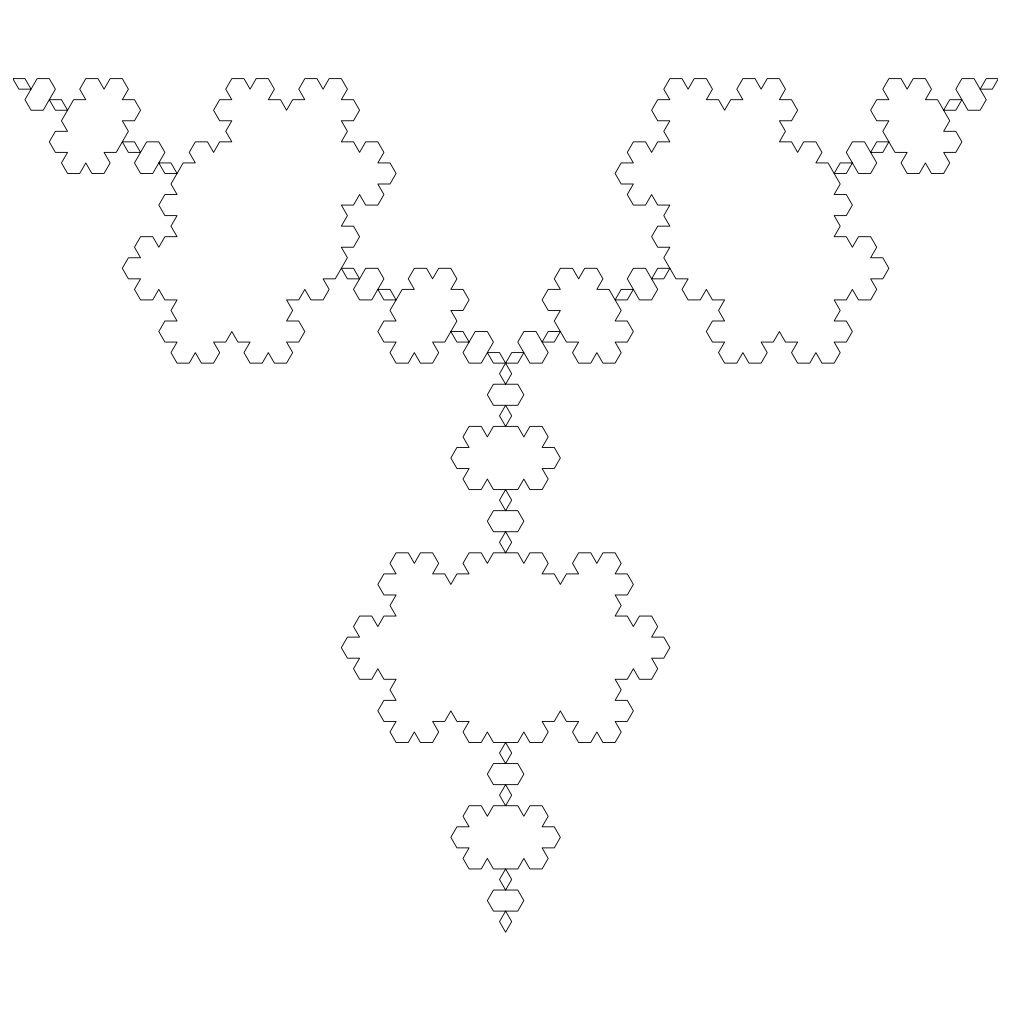}} &
        \adjustbox{valign=c}{\includegraphics[width=0.25\textwidth]{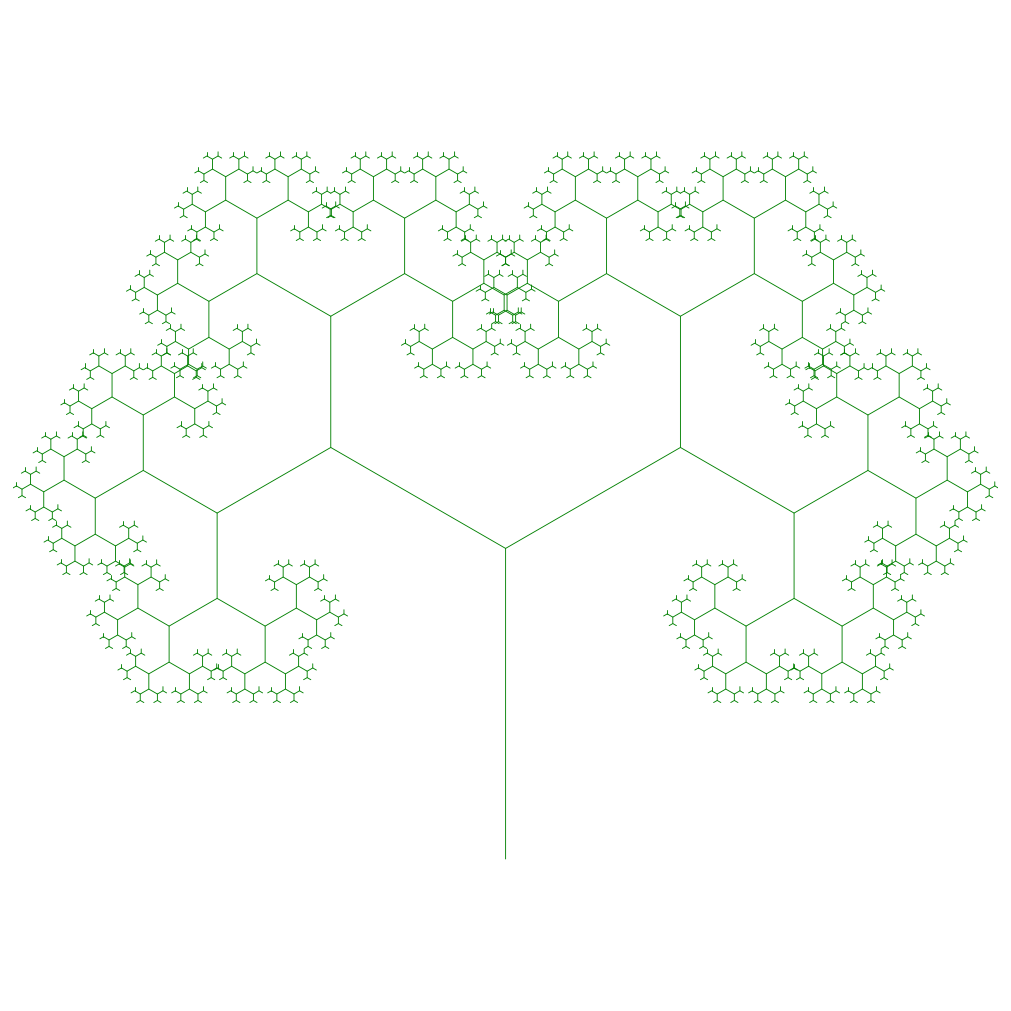}} &
        Symmetric Binary Tree \\

        \adjustbox{valign=c}{\includegraphics[width=0.25\textwidth]{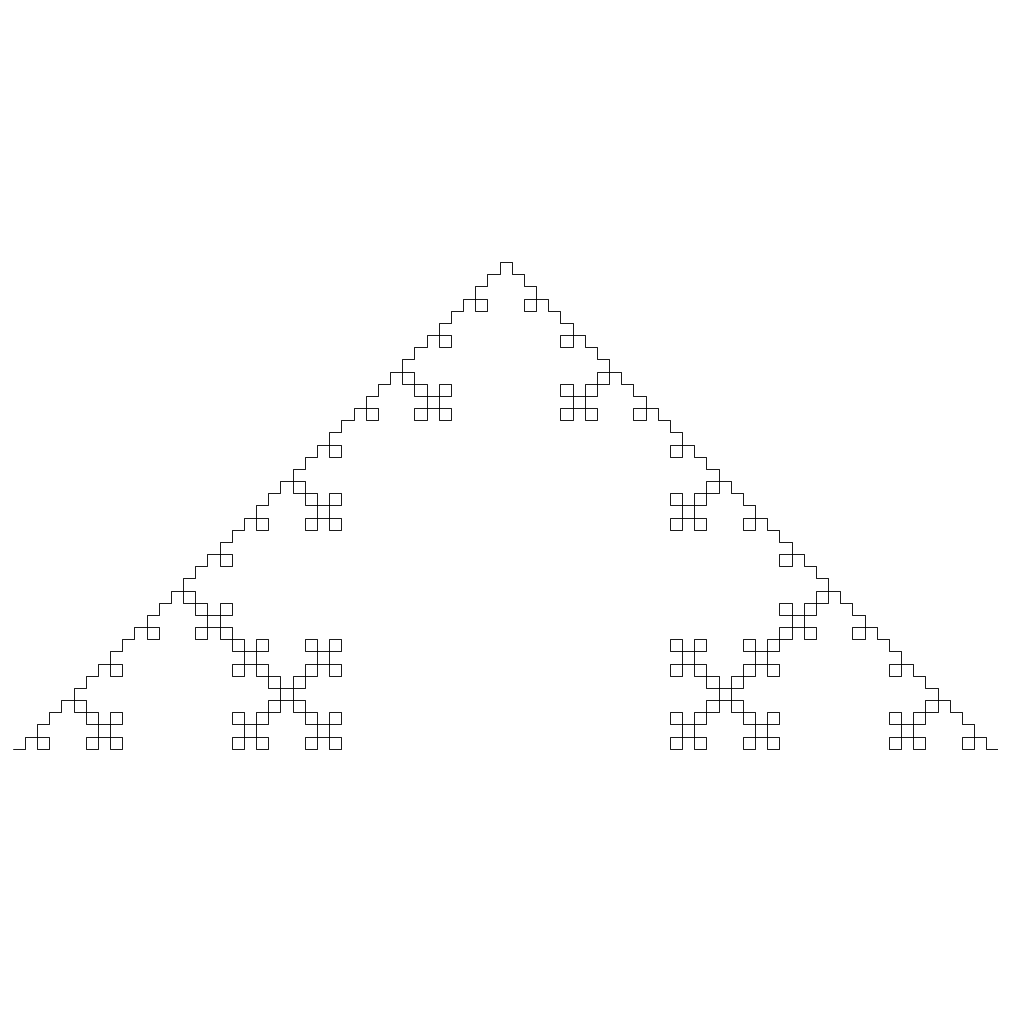}} &
        \adjustbox{valign=c}{\includegraphics[width=0.25\textwidth]{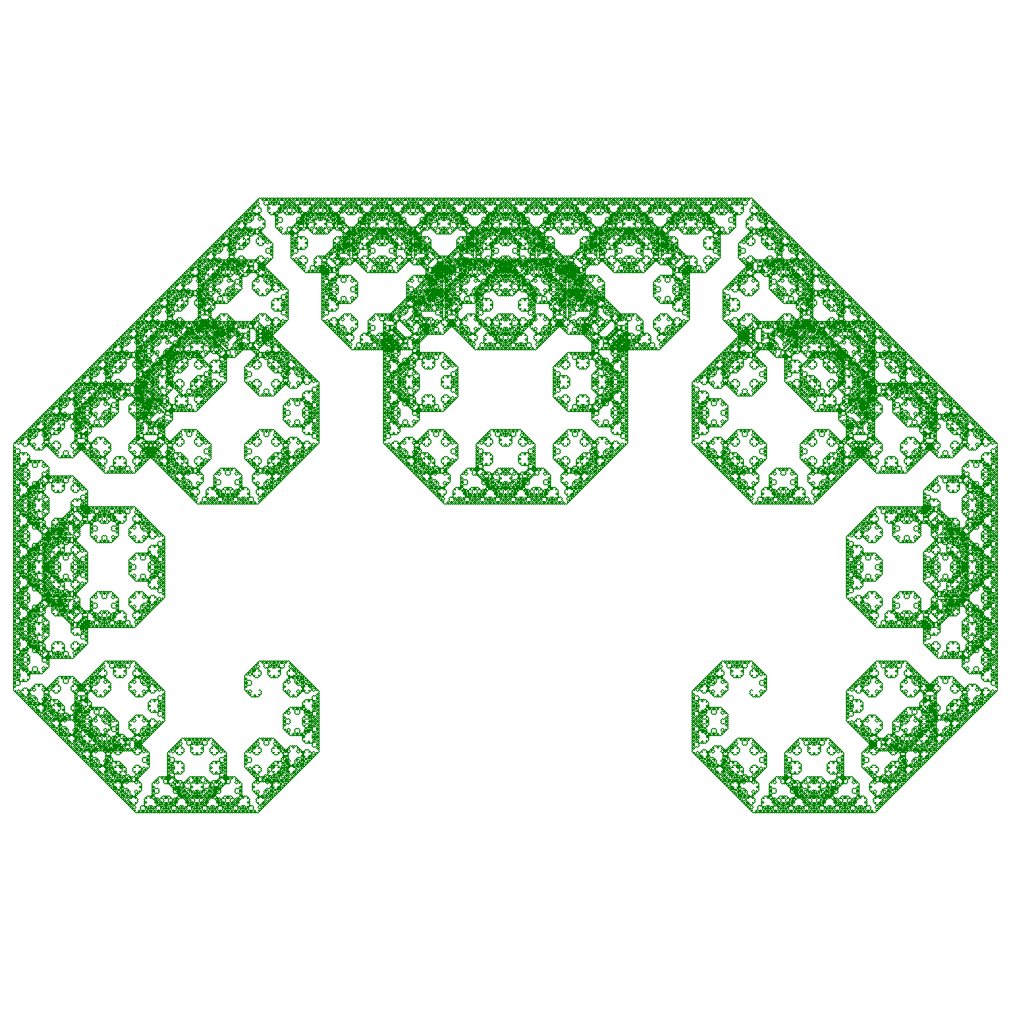}} &
        Lévy Dragon \\

        \adjustbox{valign=c}{\includegraphics[width=0.25\textwidth]{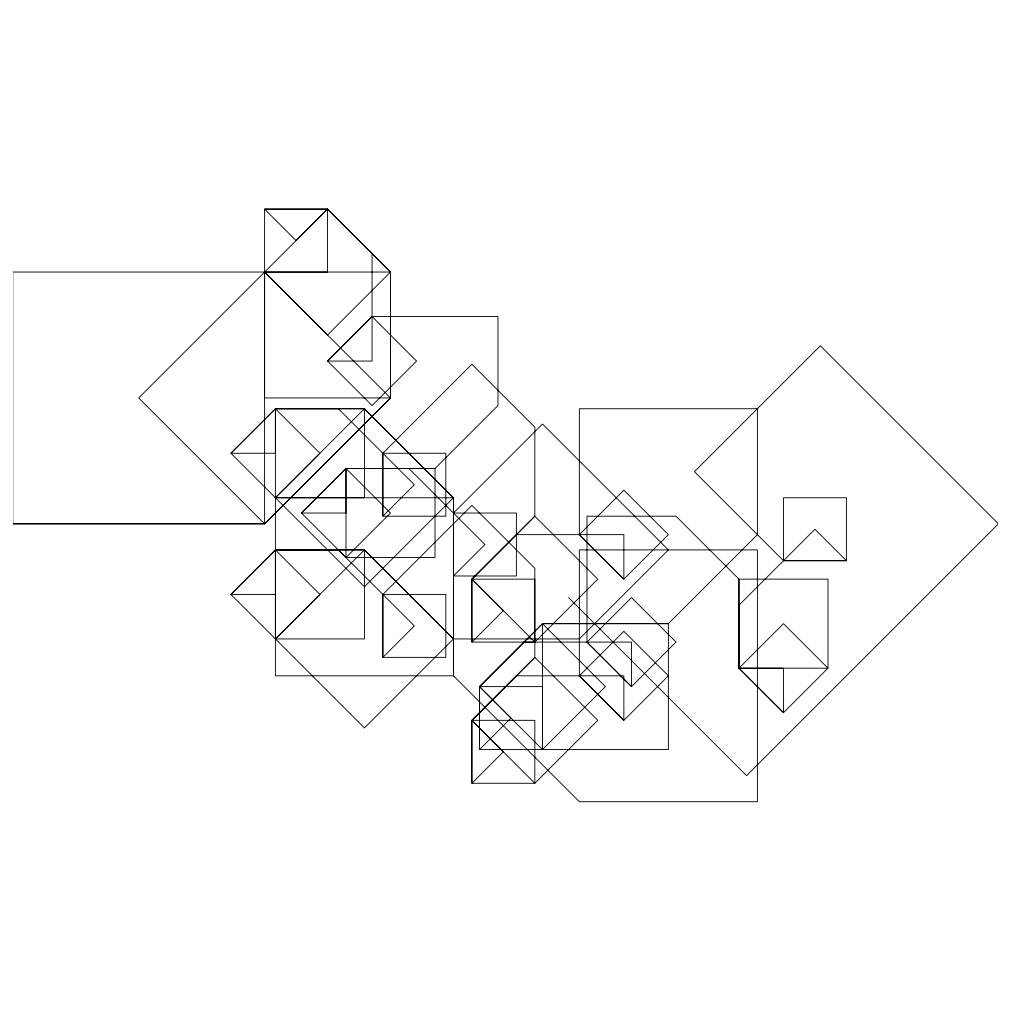}} &
        \adjustbox{valign=c}{\includegraphics[width=0.25\textwidth]{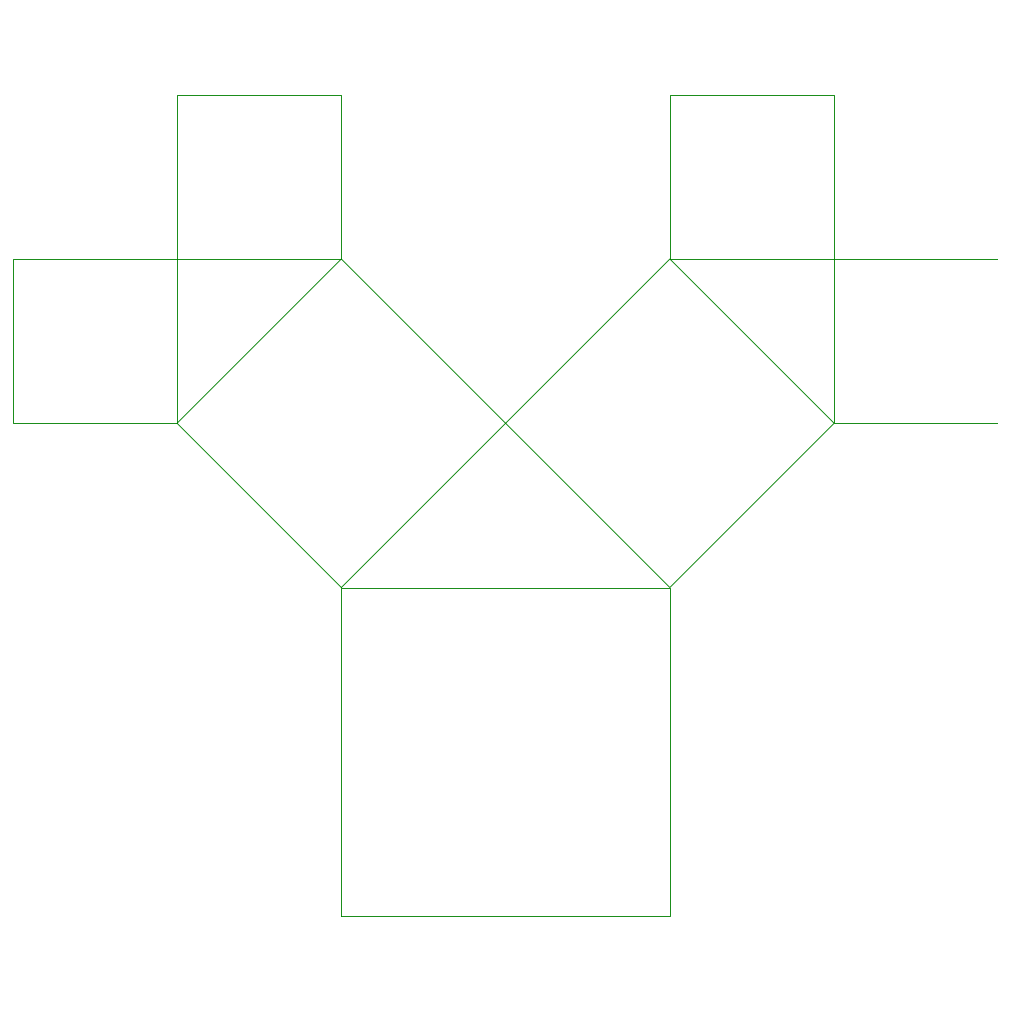}} &
        Pythagoras Tree \\

        \adjustbox{valign=c}{\includegraphics[width=0.25\textwidth]{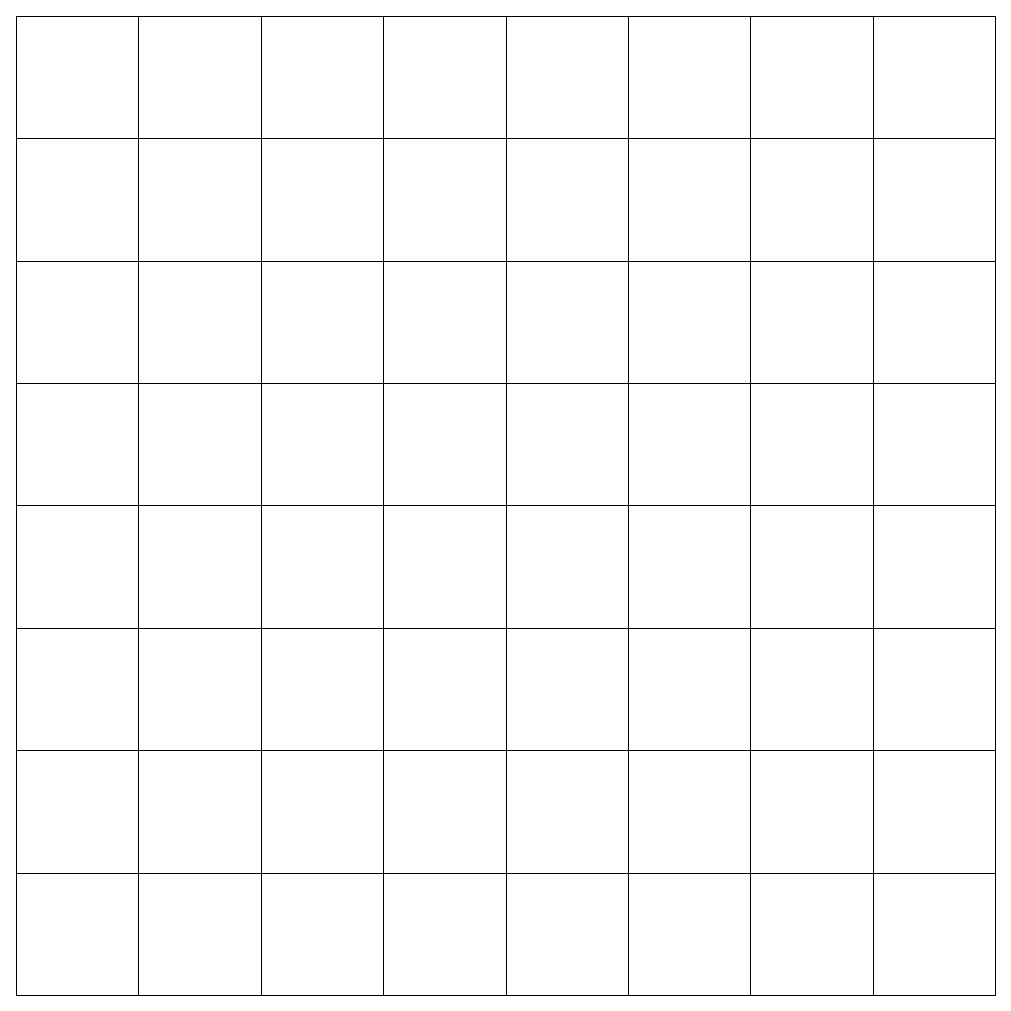}} &
        \adjustbox{valign=c}{\includegraphics[width=0.25\textwidth]{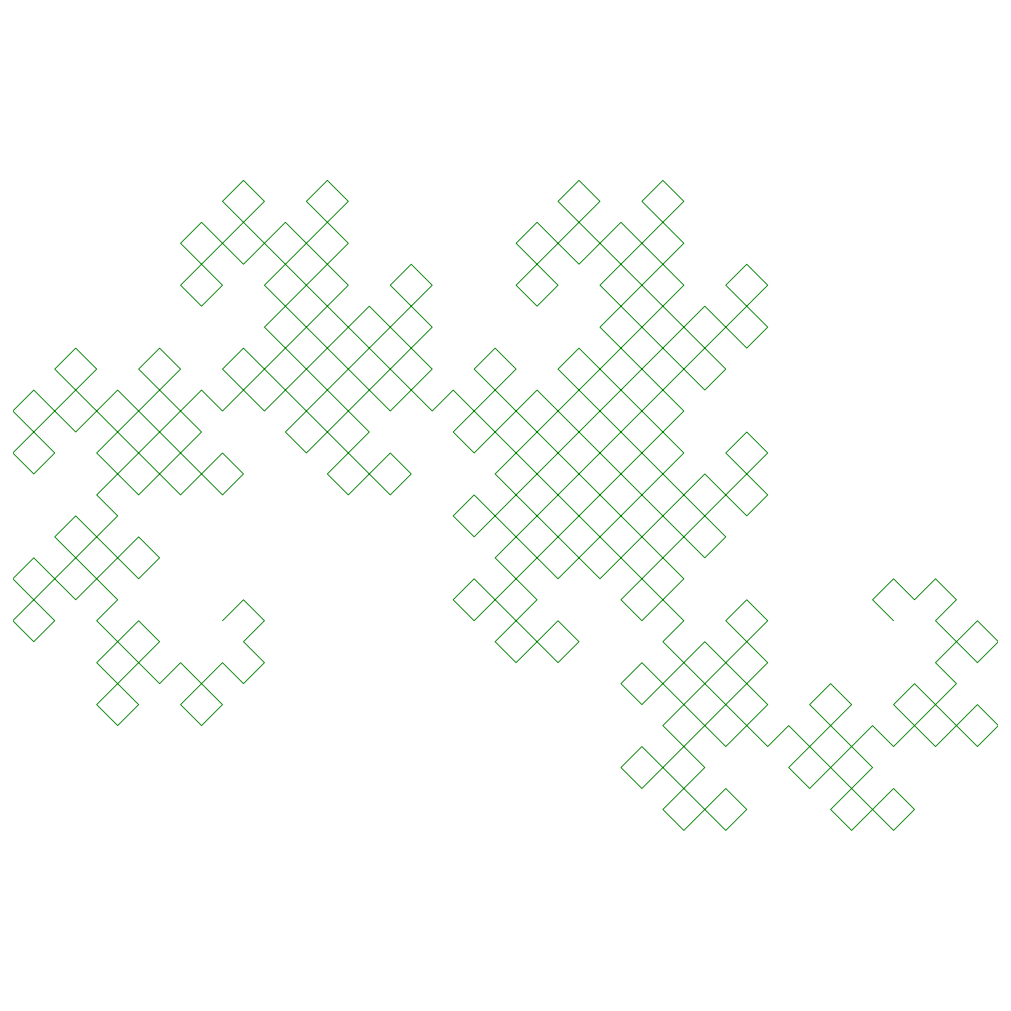}} &
        Heighway Dragon \\

        \adjustbox{valign=c}{\includegraphics[width=0.25\textwidth]{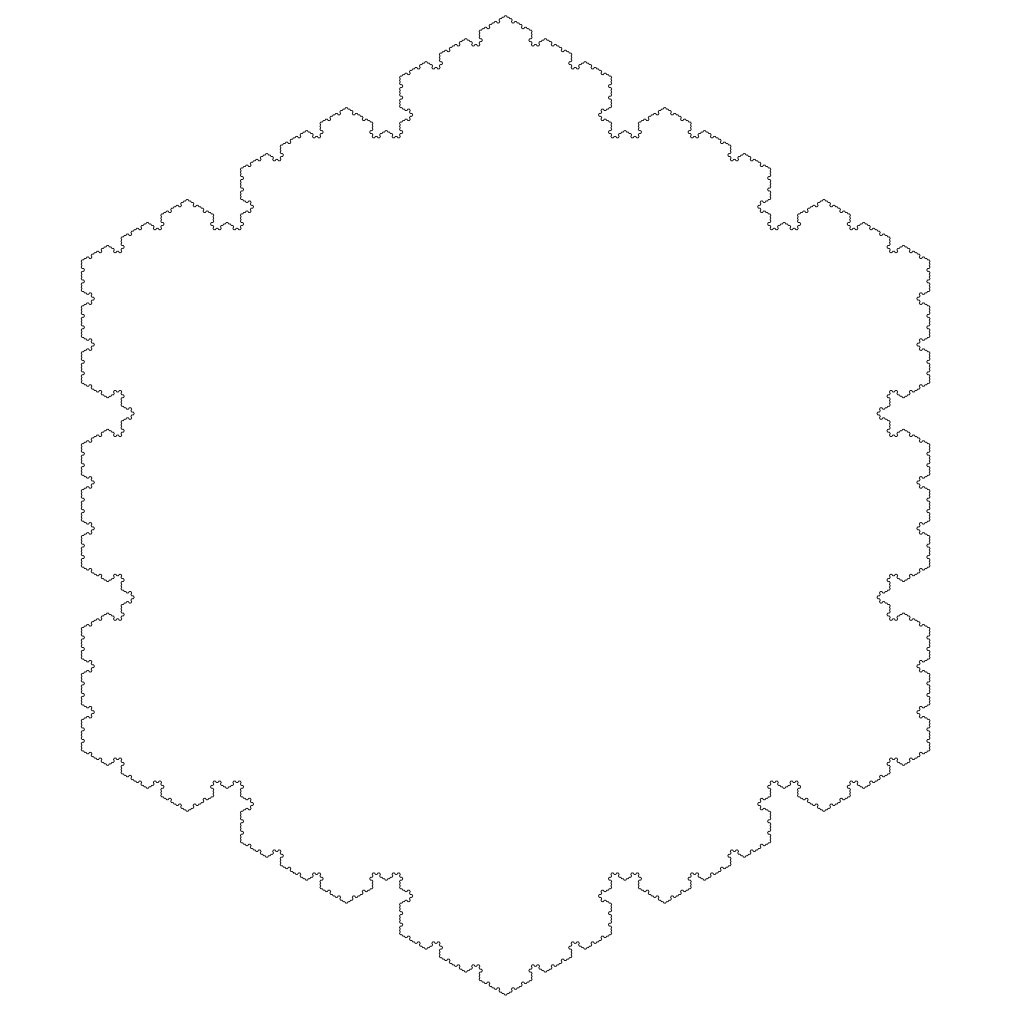}} &
        \adjustbox{valign=c}{\includegraphics[width=0.25\textwidth]{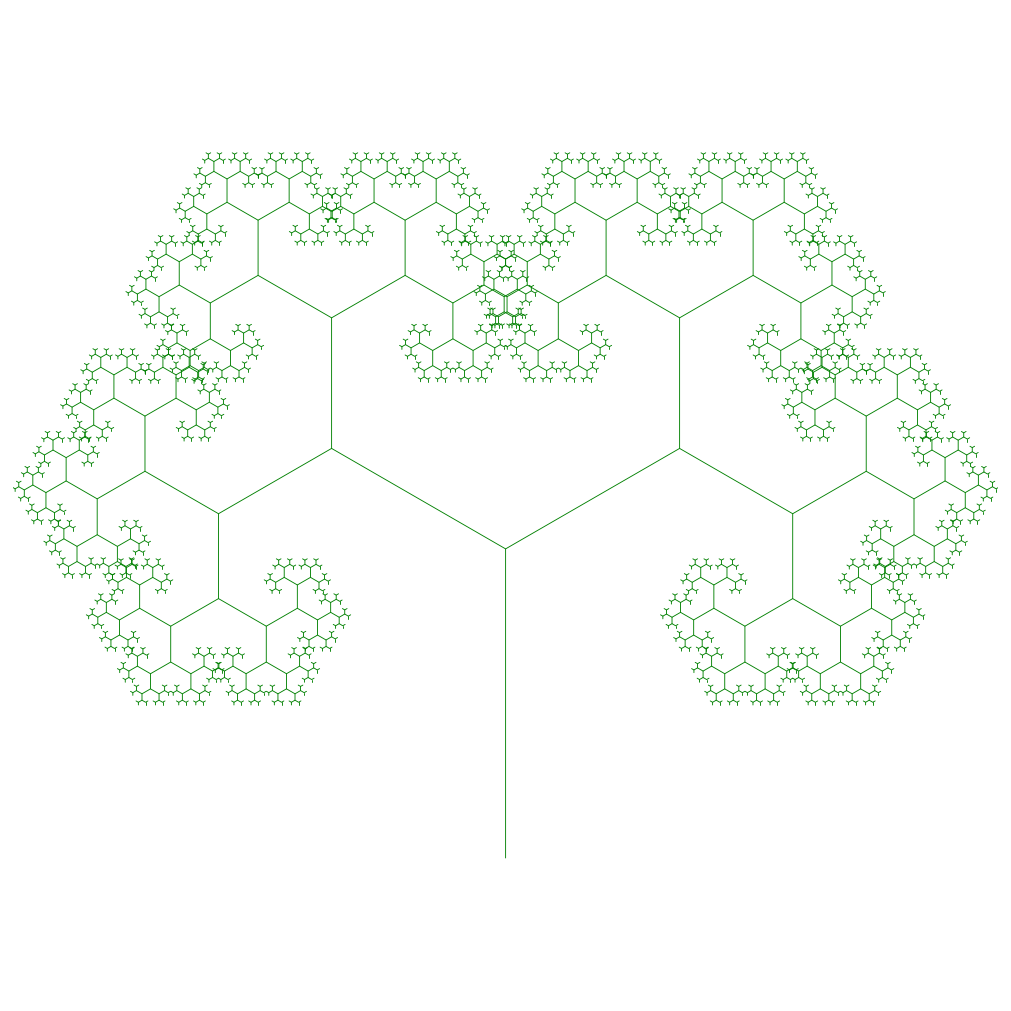}} &
        Symmetric Binary Tree
    \end{tabular}
    \caption{Representative failure cases showing model-generated green fractals (left) versus ground truth (right). These six examples achieved low similarity scores, $\textrm{IoU} \in (0.010, 0.011)$, demonstrating cases where models produced visual output but failed to implement correct fractal structures.}
    \label{fig:failed_fractals}
\end{figure}

Figure~\ref{fig:failed_fractals} shows six green fractal cases where models achieved low similarity scores (0.010-0.011) with ground truth images, representing substantial generation attempts that went wrong. Each example pairs the model's output (left) with the correct fractal (right), highlighting specific failure modes:

\begin{itemize}
    \item \textbf{Heighway Dragon (depth 8):} Model attempts to draw curved lines but fails to capture the proper recursive folding pattern and self-similarity structure.
    \item \textbf{Symmetric Binary Tree (depth 10):} Basic branching attempted but incorrect angle calculations and recursive depth handling result in malformed tree structure.
    \item \textbf{Lévy Dragon (depth 17):} Some curve generation visible but wrong turning angles and scaling factors produce unrecognizable patterns.
    \item \textbf{Pythagoras Tree (depth 2):} Square elements drawn but improper recursive placement and scaling destroy the characteristic tree-like growth pattern.
    \item \textbf{Heighway Dragon (depth 9):} Alternative dragon curve attempt showing different failure mode with incorrect segment orientations.
    \item \textbf{Symmetric Binary Tree (depth 11):} Another tree variant demonstrating systematic errors in recursive branching logic.
\end{itemize}

These failures often stem from: (1) incorrect recursive parameter propagation, (2) improper angle and scaling calculations, (3) coordinate transformation errors, (4) misunderstanding of fractal self-similarity principles, and (5) inadequate handling of recursive depth and termination conditions.

\newpage

\section{Synthesized Code Complexity Analysis}\label{sec:app:code_complexity}

One of the key motivations for evaluating fractals in the image-to-code synthesis setting is their inherent \emph{compressibility}, admitting extremely concise mathematical definitions—often specified by only a few similarity transformations in an IFS. 
From the standpoint of algorithmic information theory, this means that their Kolmogorov complexity is very low: the shortest description of the pattern is much shorter than the raw pixel-level encoding of its image.  

We thus investigate the extent to which the models capture this compactness when synthesizing code from fractal images. 
In particular, we measure the complexity of the generated programs in terms of code length. 
Shorter code is taken to reflect a more efficient internal representation of the fractal pattern, whereas longer code may suggest that the model is relying on more literal or ad-hoc descriptions rather than recognizing the recursive self-similar structure.  

Concretely, for each fractal type, we compute the number of non-blank, non-comment lines of code in the generated output. We average this length over different color instantiations of the same fractal to reduce variance, and track how it evolves with the fractal recursion depth. This metric gives an empirical proxy for the compactness of the synthesized code.  

\begin{figure}[h!]
\centering
\includegraphics[width=\textwidth]{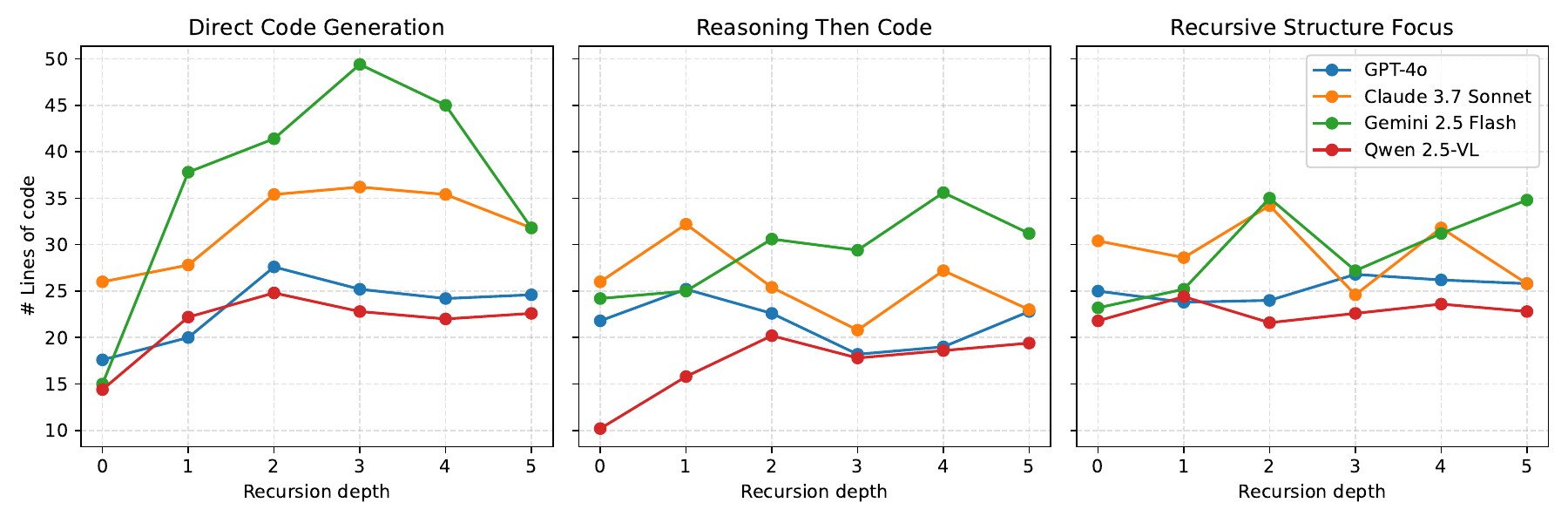}\vspace{-8mm}
\caption{\textbf{Cantor Set:} Synthesized code complexity (\# non-blank, non-comment lines of code, averaged over colors) vs.~recursion depth comparing all models for each prompting strategy.}\vspace{-4mm}
\label{fig:code_complexity_cantor_set}
\end{figure}

\begin{figure}[h!]
\centering
\includegraphics[width=\textwidth]{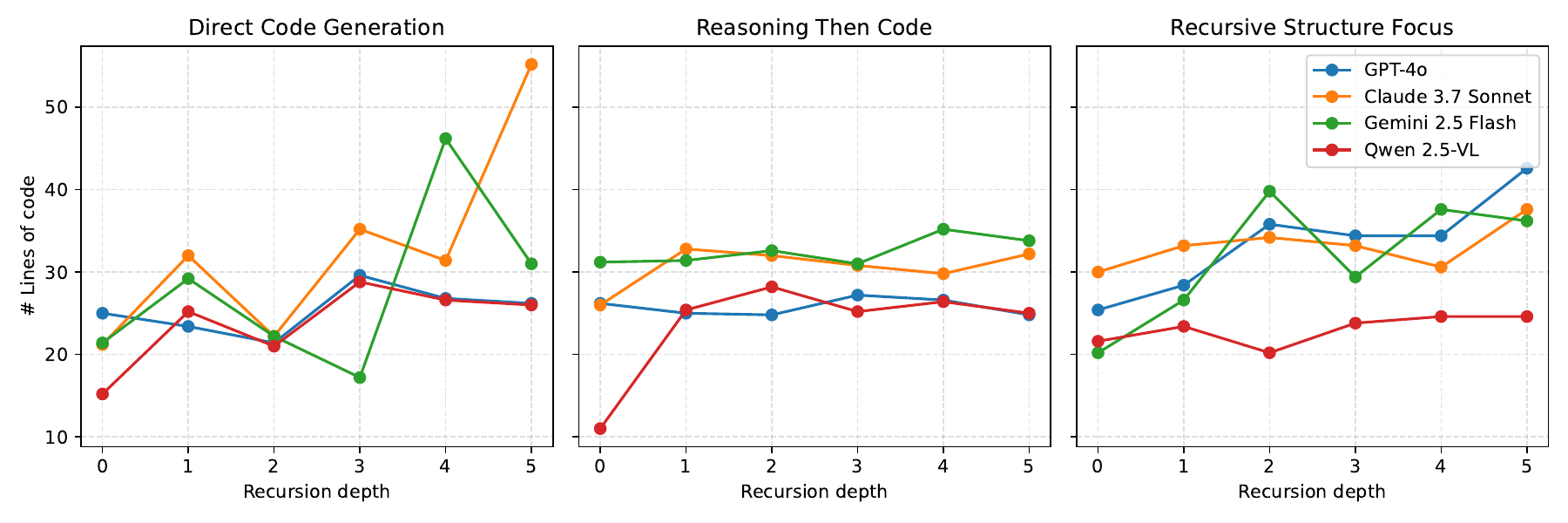}\vspace{-8mm}
\caption{\textbf{Cantor Dust:} Synthesized code complexity (\# non-blank, non-comment lines of code, averaged over colors) vs.~recursion depth comparing all models for each prompting strategy.}\vspace{-4mm}
\label{fig:code_complexity_cantor_dust}
\end{figure}

\begin{figure}[h!]
\centering
\includegraphics[width=\textwidth]{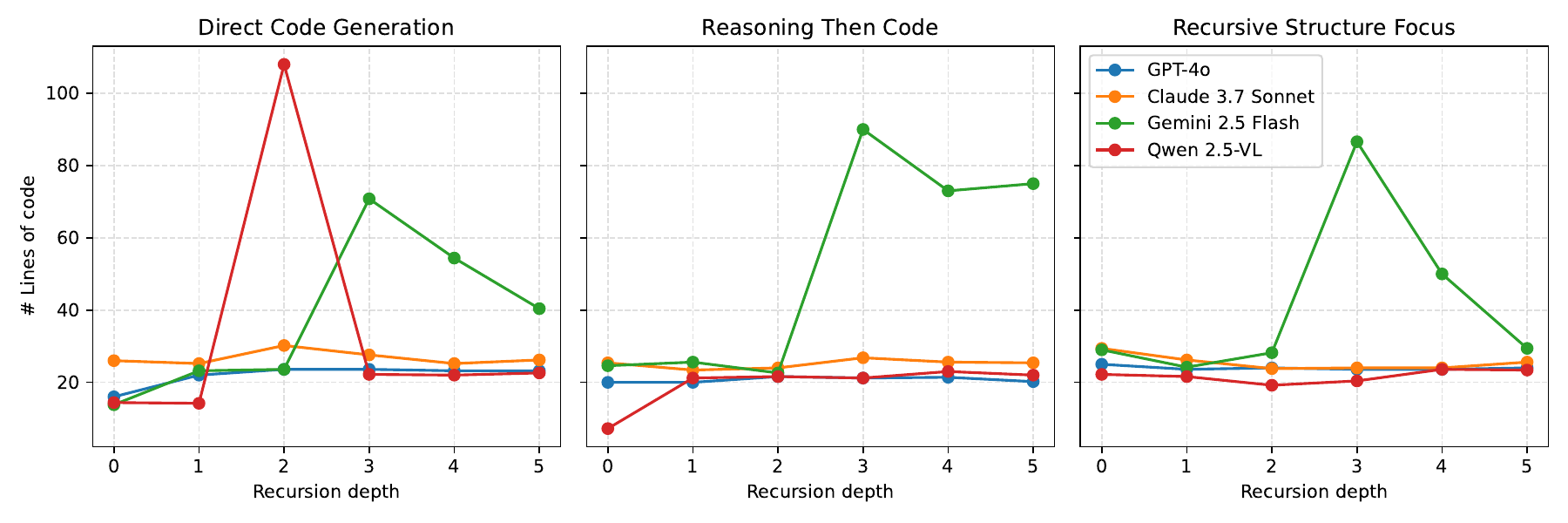}\vspace{-8mm}
\caption{\textbf{Koch Curve:} Synthesized code complexity (\# non-blank, non-comment lines of code, averaged over colors) vs.~recursion depth comparing all models for each prompting strategy.}\vspace{-4mm}
\label{fig:code_complexity_koch_curve}
\end{figure}

\begin{figure}[h!]
\centering
\includegraphics[width=\textwidth]{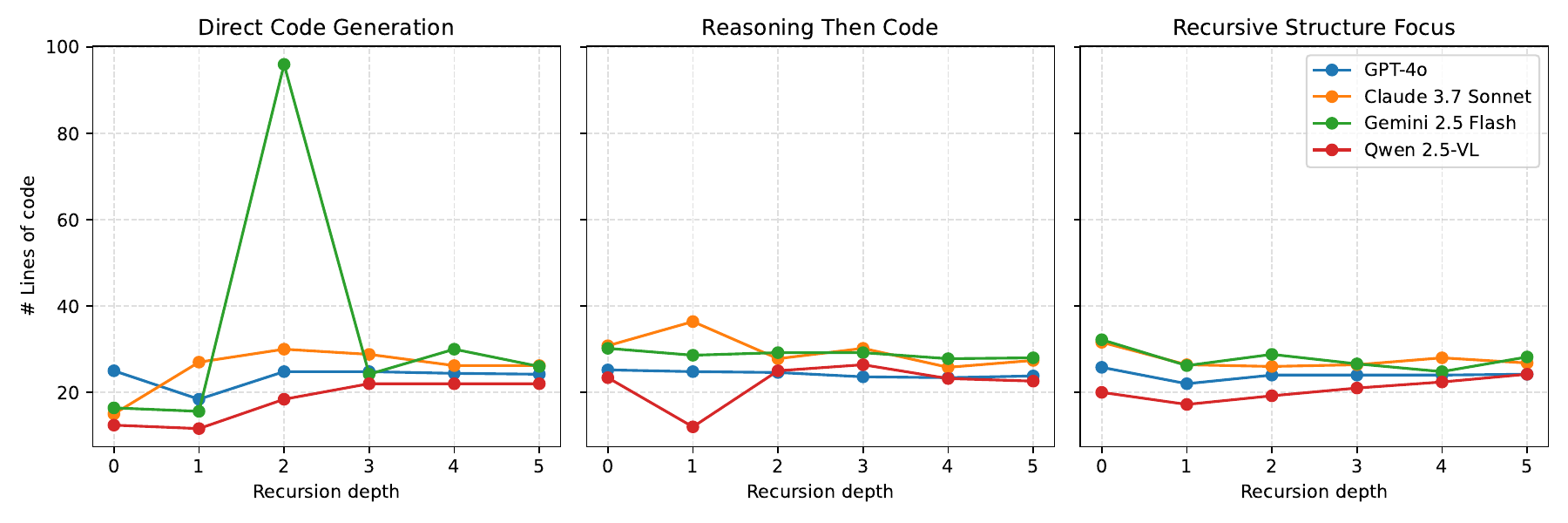}\vspace{-8mm}
\caption{\textbf{Koch Snowflake:} Synthesized code complexity (\# non-blank, non-comment lines of code, averaged over colors) vs.~recursion depth comparing all models for each prompting strategy.}\vspace{-4mm}
\label{fig:code_complexity_koch_snowflake}
\end{figure}

\begin{figure}[h!]
\centering
\includegraphics[width=\textwidth]{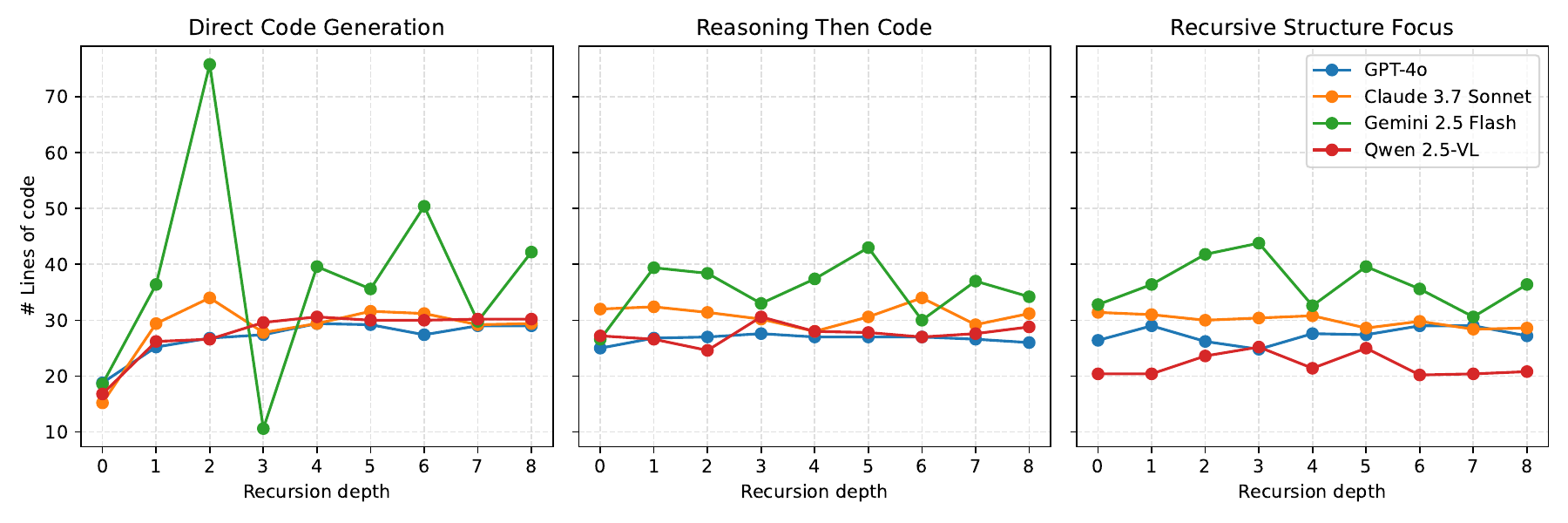}\vspace{-8mm}
\caption{\textbf{Sierpiński Gasket:} Synthesized code complexity (\# non-blank, non-comment lines of code, averaged over colors) vs.~recursion depth comparing all models for each prompting strategy.}\vspace{-4mm}
\label{fig:code_complexity_sierpinski_gasket}
\end{figure}

\begin{figure}[h!]
\centering
\includegraphics[width=\textwidth]{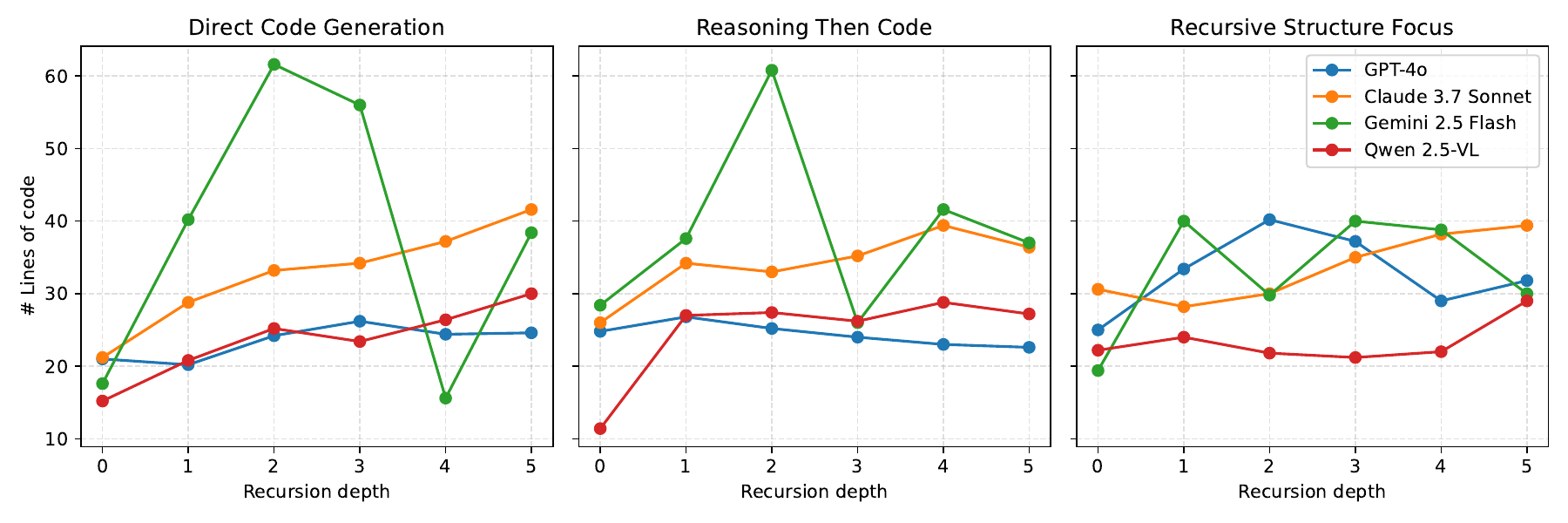}\vspace{-8mm}
\caption{\textbf{Sierpiński Carpet:} Synthesized code complexity (\# non-blank, non-comment lines of code, averaged over colors) vs.~recursion depth comparing all models for each prompting strategy.}\vspace{-4mm}
\label{fig:code_complexity_sierpinski_carpet}
\end{figure}

\begin{figure}[h!]
\centering
\includegraphics[width=\textwidth]{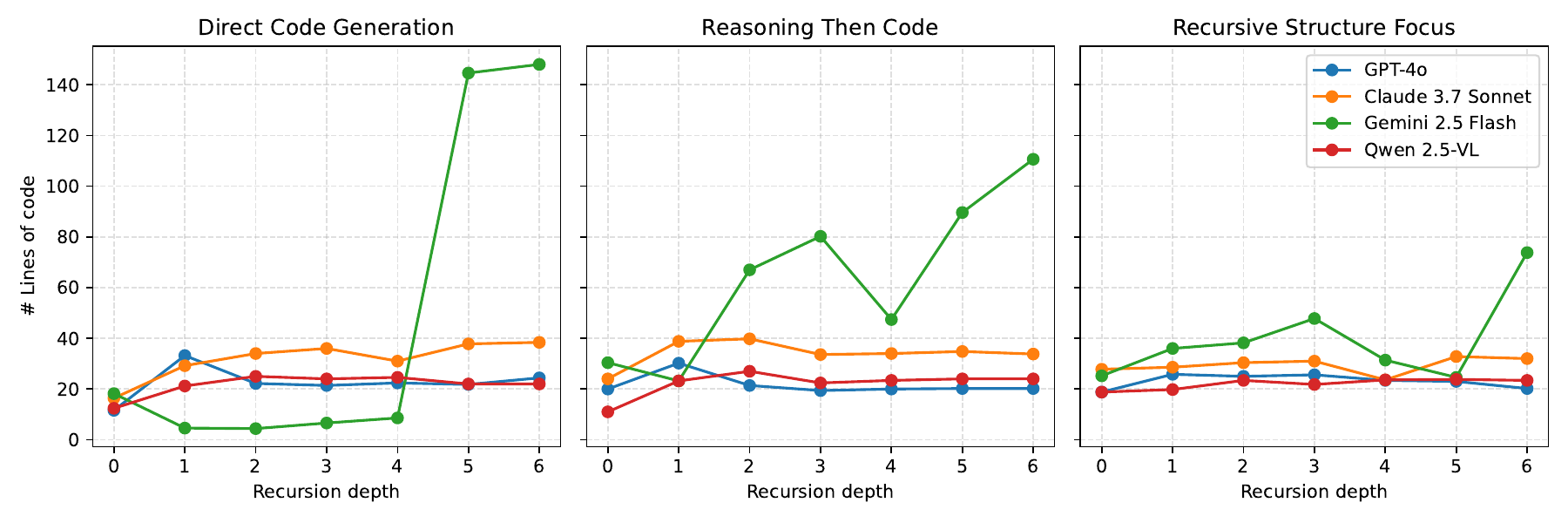}\vspace{-8mm}
\caption{\textbf{Sierpiński Pentagon:} Synthesized code complexity (\# non-blank, non-comment lines of code, averaged over colors) vs.~recursion depth comparing all models for each prompting strategy.}\vspace{-4mm}
\label{fig:code_complexity_sierpinski_pentagon}
\end{figure}

\begin{figure}[h!]
\centering
\includegraphics[width=\textwidth]{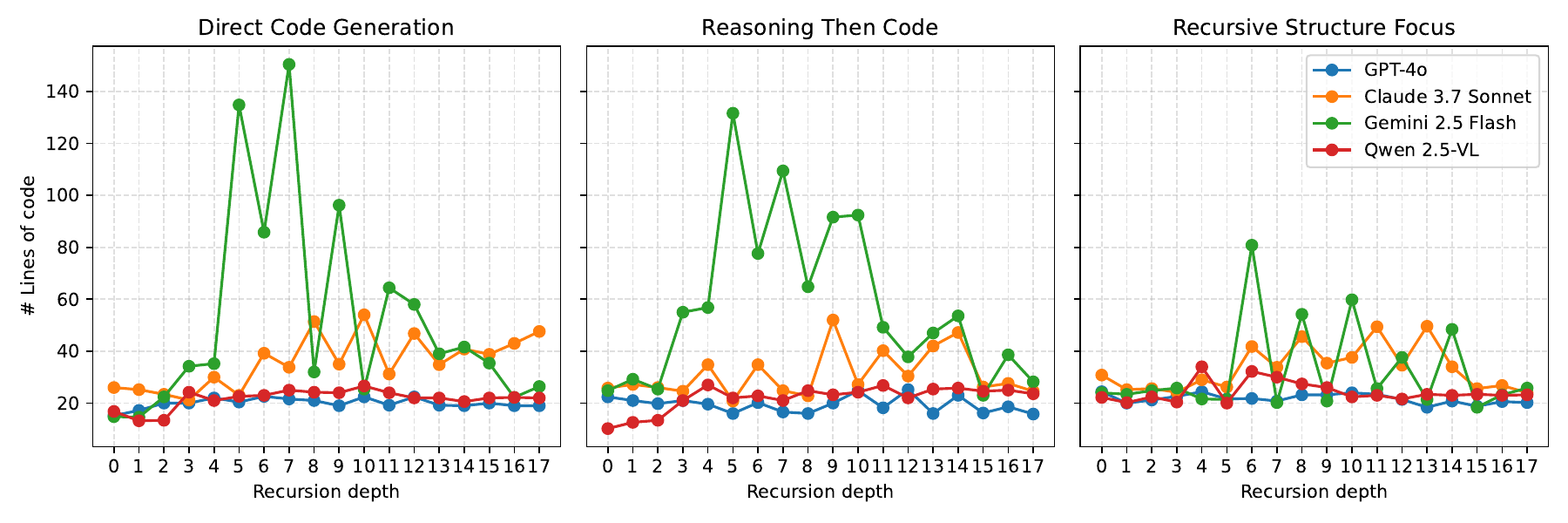}\vspace{-8mm}
\caption{\textbf{Heighway Dragon:} Synthesized code complexity (\# non-blank, non-comment lines of code, averaged over colors) vs.~recursion depth, comparing all models for each prompting strategy.}\vspace{-4mm}
\label{fig:code_complexity_heighway_dragon}
\end{figure}

\begin{figure}[h!]
\centering
\includegraphics[width=\textwidth]{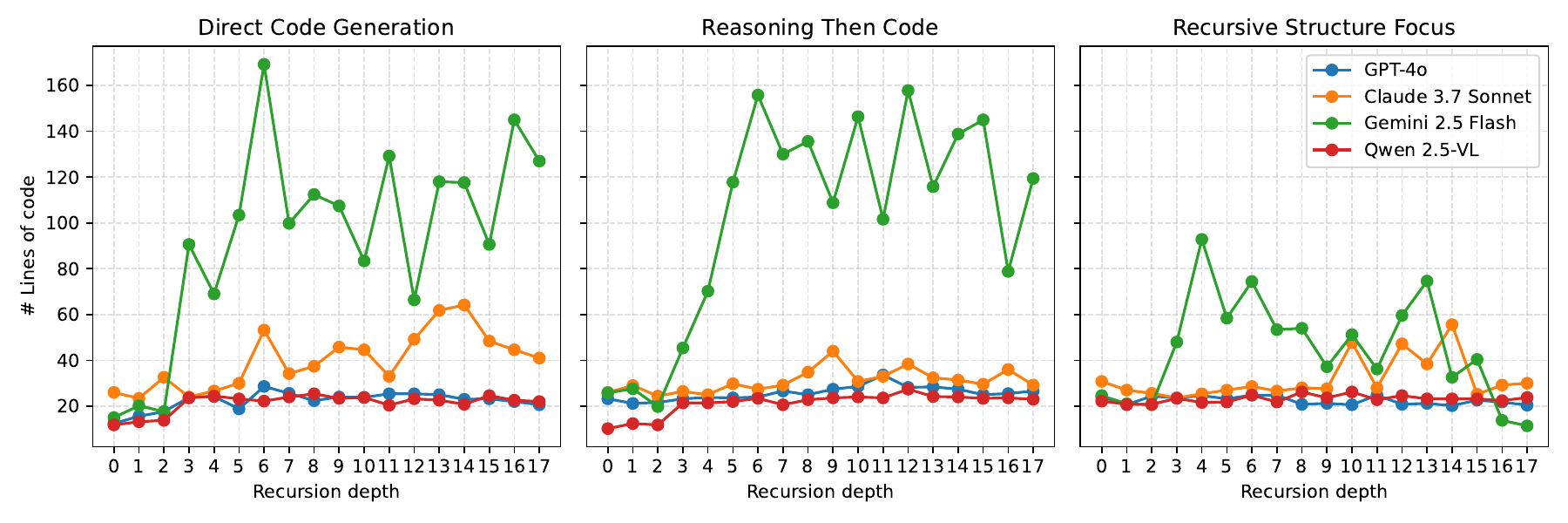}\vspace{-8mm}
\caption{\textbf{Lévy Dragon:} Synthesized code complexity (\# non-blank, non-comment lines of code, averaged over colors) vs.~recursion depth comparing all models for each prompting strategy.}\vspace{-4mm}
\label{fig:code_complexity_levy_dragon}
\end{figure}

\begin{figure}[h!]
\centering
\includegraphics[width=\textwidth]{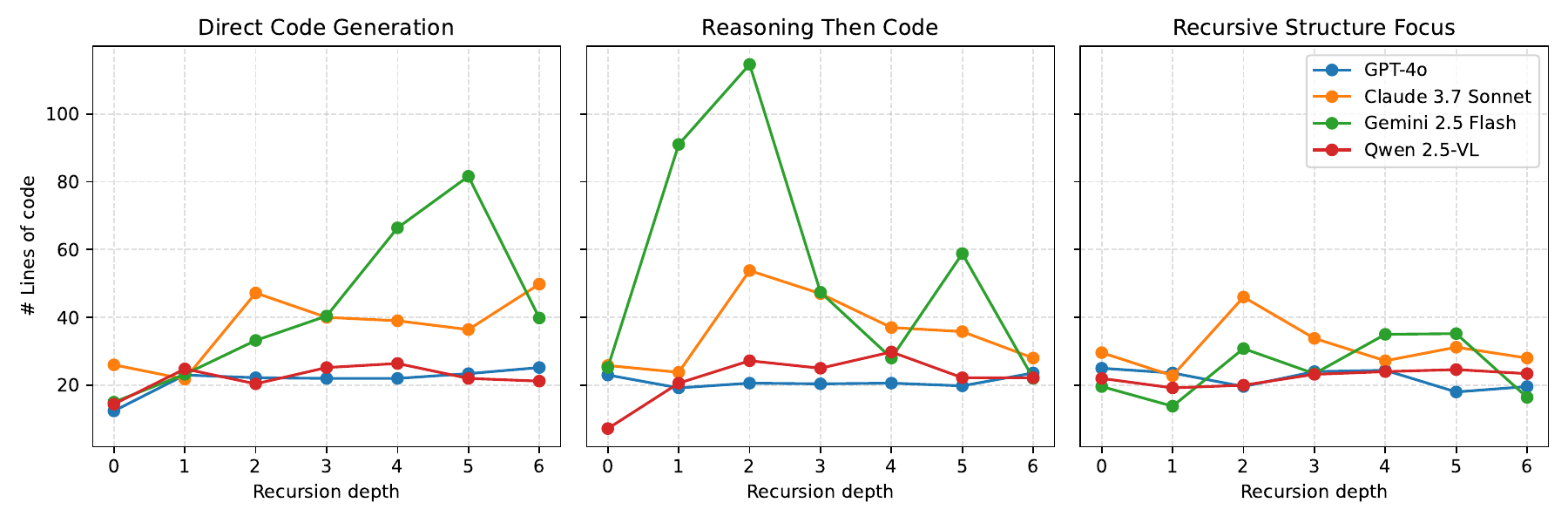}\vspace{-8mm}
\caption{\textbf{McWorter's Pentigree:} Synthesized code complexity (\# non-blank, non-comment lines of code, averaged over colors) vs.~recursion depth comparing all models for each prompting strategy.}\vspace{-4mm}
\label{fig:code_complexity_mcworter_pentigree}
\end{figure}

\begin{figure}[h!]
\centering
\includegraphics[width=\textwidth]{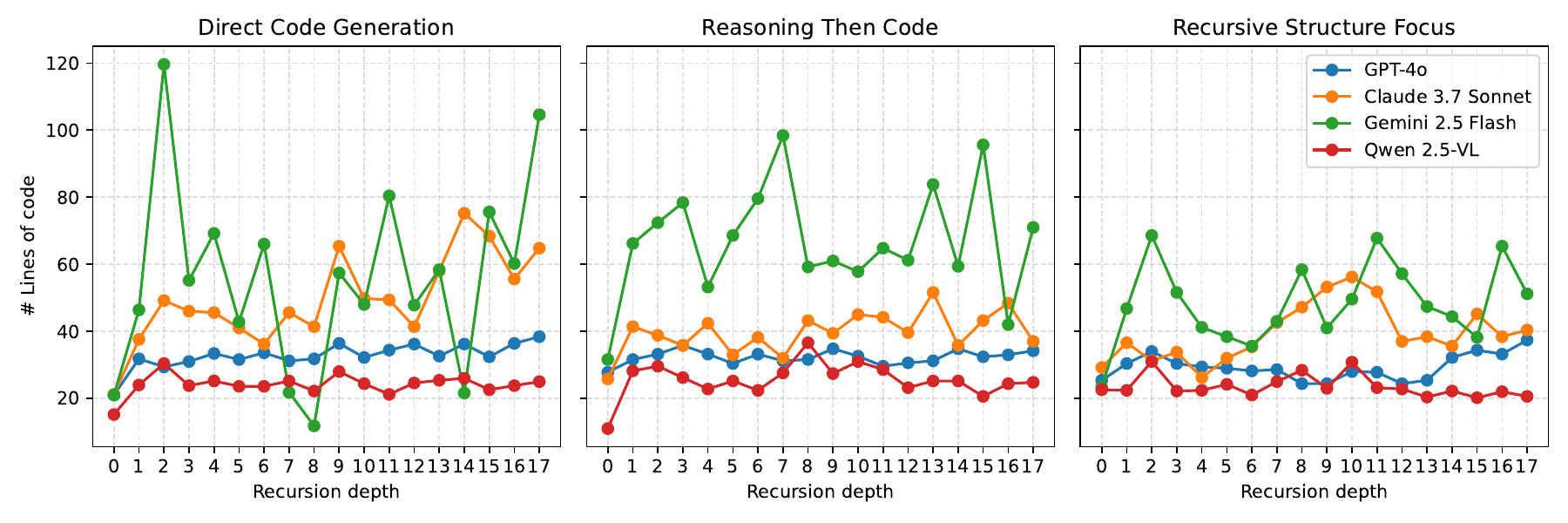}\vspace{-8mm}
\caption{\textbf{Pythagoras Tree:} Synthesized code complexity (\# non-blank, non-comment lines of code, averaged over colors) vs.~recursion depth comparing all models for each prompting strategy.}\vspace{-4mm}
\label{fig:code_complexity_pythagoras_tree}
\end{figure}

\begin{figure}[h!]
\centering
\includegraphics[width=\textwidth]{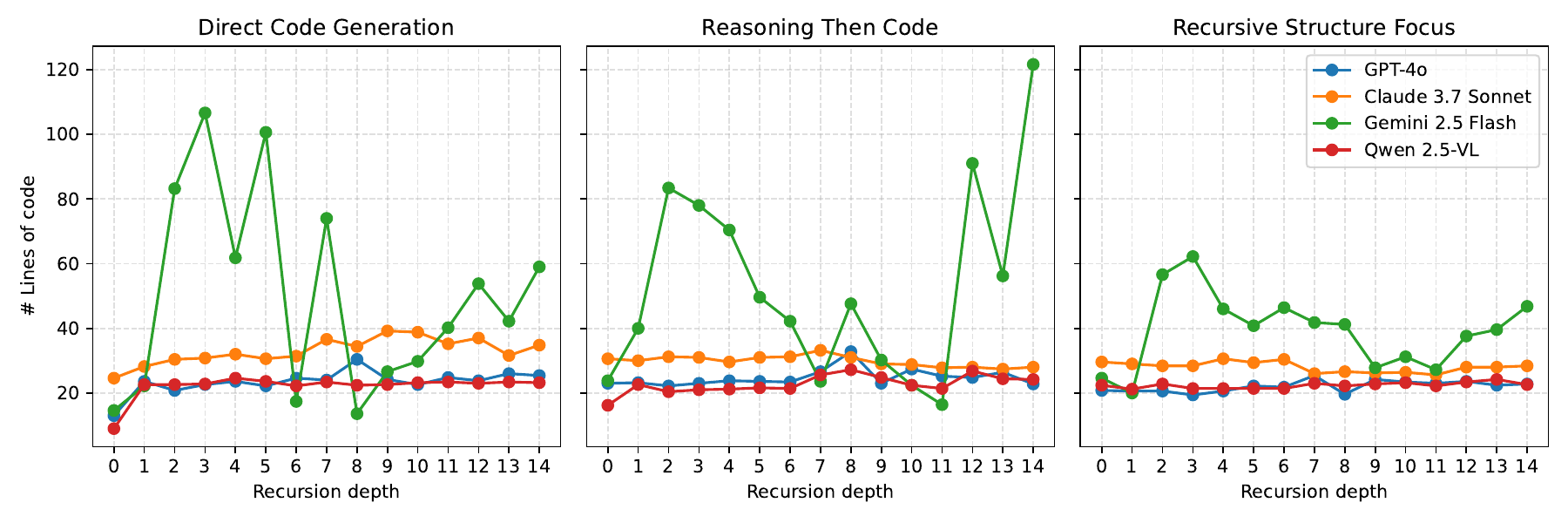}\vspace{-8mm}
\caption{\textbf{Symmetric Binary Tree:} Synthesized code complexity (\# non-blank, non-comment lines of code, averaged over colors) vs.~recursion depth comparing all models for each prompting strategy.}\vspace{-4mm}
\label{fig:code_complexity_symmetric_binary_tree}
\end{figure}

Figures~\ref{fig:code_complexity_cantor_set}--\ref{fig:code_complexity_symmetric_binary_tree} compare code complexity across fractals and prompting strategies.
Overall, Gemini produces consistently more verbose code than the other models.
In several cases—most notably the Heighway Dragon—we observe a phase-transition–like behavior: complexity grows with recursion depth up to a threshold, after which the model appears to recognize and exploit the recursive structure, yielding a sharp reduction in code length.
Prompts that explicitly emphasize recursion tend to reduce complexity, whereas prompts encouraging step-by-step reasoning show little measurable impact.

\clearpage

\clearpage
\section{Extended Discussion of Limitations}\label{sec:app:limitations}

\subsection{Evaluation Methodology}\label{sec:app:limitations_eval}

\paragraph{Single Generation and Stochasticity.}
We perform a single generation per image, which may not fully capture model stochasticity. Observed color-level fluctuations in some cases (e.g., Gemini 2.5 Flash: 11.5\% black vs 4.1\% purple in Tab.~\ref{tab:comprehensive_overview}) likely stem from sampling variance rather than genuine difficulty differences, as empirically Tab.~\ref{tab:color_analysis} shows <1.5\% average difference across colors. Averaging multiple generations per image would improve statistical reliability and enable rigorous significance testing with confidence intervals and variance analysis.

\paragraph{Pixel-Based Similarity Limitations.}
We use pixel-based similarity (IoU $\ge$ 95\%) as the primary correctness criterion, following similar vision-to-code benchmarks such as TurtleBench~\cite{rismanchian-etal-2025-turtlebench}.
While effective for judging visual fidelity, this binary threshold does not assess how closely a model captures the underlying generative structure.
Minor errors in geometric parameters—such as small angle deviations—can compound recursively, producing large visual discrepancies even when the intended rule is nearly correct.
Incorporating multi-threshold or continuous similarity analysis would yield finer diagnostic resolution.

\paragraph{Structure-Aware Metrics.}
Complementary \emph{structure-aware} metrics—such as branch count accuracy, angle set matching, recursive depth detection, or IFS parameter extraction—could further provide finer-grained diagnostics of which specific geometric or recursive properties models capture versus miss.
Such metrics would distinguish, for example, between code that produces visually similar output through incorrect means (e.g., iterative approximation) versus code that implements the correct generative process with minor visual artifacts.

\subsection{Benchmark Scope}\label{sec:app:limitations_scope}

\paragraph{Program Synthesis Baselines.}
\FB{} is designed as a diagnostic benchmark for current multimodal large language models (MLLMs), focusing on vision-to-code reasoning rather than full program synthesis pipelines.
We therefore do not include comparisons with traditional program synthesis baselines such as symbolic search, neurosymbolic inference, and constraint-based synthesis~\cite{chaudhuri2025neurosymbolic,gulwani2017program}.
Such comparisons would, however, provide useful performance context, situating MLLM capabilities relative to specialized synthesis techniques.

\paragraph{Model Coverage.}
Our evaluation covers four representative MLLMs available at the time of benchmark development (GPT-4o, Claude 3.7 Sonnet, Gemini 2.5 Flash, and Qwen 2.5-VL). More recent reasoning-specialized models (e.g., OpenAI o1, DeepSeek-R1) or math-focused models are not included. Evaluating such models would help determine whether \FB{} distinguishes genuinely stronger reasoning capabilities or primarily exposes universal failure modes in visual-mathematical abstraction.
Expanding coverage to these systems would sharpen the benchmark's diagnostic scope.

\subsection{Analysis Limitations}\label{sec:app:limitations_analysis}

\paragraph{Observational Findings.}
The observed differences in prompting strategies—where direct code generation unexpectedly outperforms reasoning-first prompts (Sec.~\ref{sec:prompting_strategy_analysis})—remain observational.
We do not perform stepwise ablations to isolate whether the performance gap stems from prompt complexity overload, genuine incompatibility between verbal reasoning and geometric precision, or other factors. Controlled experiments varying prompt length, reasoning depth, and instruction complexity independently would enable causal interpretation beyond our current hypotheses.

\paragraph{Model Improvement Feedback Loop.}
While \FB{} systematically identifies characteristic failure modes in recursive and branching reasoning, we do not demonstrate how these diagnostics could guide targeted model improvements through few-shot tuning, structured prompting refinements, or tool integration.
Establishing such a feedback loop—where benchmark insights lead to measurable capability gains—would further validate \FB{} as a tool for advancing visual–mathematical reasoning research.


\end{document}